\documentclass[12pt]{article} %
\usepackage{times}
\usepackage{url}            %
\usepackage{booktabs}       %
\usepackage{amsfonts}       %
\usepackage{multirow}

\usepackage{tikz}
\usetikzlibrary{arrows}
\usepackage{thmtools}
\usepackage{booktabs}
\usepackage{nicefrac}       %
\usepackage{microtype}      %
\usepackage{xspace}
\usepackage{multirow}
\usepackage{booktabs}
\usepackage{amsmath}
\usepackage{mkolar_definitions}

\usepackage{algorithm}
\usepackage{algorithmic}
\usepackage{color}

\usepackage{authblk}
\usepackage{color, colortbl}
\usepackage{enumitem}
\usepackage{comment}
\usepackage{bm}
\usepackage{subcaption}
\usepackage[margin=1.0in]{geometry}
\usepackage{amsbsy}
\usepackage{wrapfig}
\usepackage{thmtools}

\usepackage{url}
\usepackage{authblk}
\usepackage{csquotes}
\usepackage{tikz}

\usepackage[colorlinks=true, linkcolor=blue, citecolor=blue]{hyperref}

\usepackage{natbib}

\usepackage{centernot}

\newcount\Comments  %
\definecolor{darkgreen}{rgb}{0,0.5,0}
\definecolor{darkred}{rgb}{0.7,0,0}
\definecolor{teal}{rgb}{0.3,0.8,0.8}
\definecolor{orange}{rgb}{1.0,0.5,0.0}
\definecolor{purple}{rgb}{0.8,0.0,0.8}
\newcommand{\kibitz}[2]{\ifnum\Comments=1{\textcolor{#1}{\textsf{\footnotesize #2}}}\fi}
\usetikzlibrary{positioning}

\definecolor{Gray}{gray}{0.9}

\newcommand{\KL}{\mathrm{KL}}

\newcommand{\data}{\mathrm{data}}

\newcommand{\pre}{\mathrm{pre}}

\newcommand{\newedit}{\color{purple}}
\newcommand{\rI}{\mathrm{I}}

\definecolor{columbiablue}{rgb}{0.61, 0.87, 1.0}

\newenvironment{keyword}[1] %
{
    \par\noindent\textbf{#1:} %
    \itshape %
}
{
    \par\normalfont %
}

\usepackage{colortbl}
\definecolor{Gray}{gray}{0.9}
\newcolumntype{g}{>{\columncolor{Gray}}c}

\usepackage[most,skins,theorems]{tcolorbox}

\tcbset{
  aibox/.style={
    width=\linewidth,
    top=8pt,
    bottom=4pt,
    colback=red!6!white,
    colframe=black,
    colbacktitle=pink,
    enhanced,
    center,
    attach boxed title to top left={yshift=-0.1in,xshift=0.15in},
    boxed title style={boxrule=0pt,colframe=white},
  }
}
\newtcolorbox{AIbox}[2][]{aibox,title=#2,#1}

\tcbset{
  Thmbox/.style={
    width=\linewidth,
    top=2pt,
    bottom=2pt,
    colback=purple!6!white,
    colframe=black,
    enhanced,
    center,
  }
}
\newtcolorbox{Thmbox}[2][]{Thmbox}

\tcbset{
  Exabox/.style={
    width=\linewidth,
    top=2pt,
    bottom=2pt,
    colback=green!6!white,
    colframe=black,
    enhanced,
    center,
  }
}
\newtcolorbox{Exabox}[2][]{Exabox}

\ifdefined\usebigfont

\usepackage{times}
\usepackage[fontsize=13pt]{scrextend}
\AtBeginDocument{\newgeometry{letterpaper,left=0.56in,right=0.56in,top=1.71in,bottoj=1.77in}}

\else
\fi

\title{Inference-Time Alignment in Diffusion Models with Reward-Guided Generation: Tutorial and Review}

\author{
    Masatoshi Uehara \textsuperscript{1}\thanks{ueharamasatoshi136@gmail.com}, 
  Yulai Zhao \textsuperscript{2},
 Chenyu Wang \textsuperscript{3},
 Xiner Li \textsuperscript{4},  \\ 
 Aviv Regev \textsuperscript{1}, 
 Sergey Levine \textsuperscript{5$*$} , 
Tommaso Biancalani \textsuperscript{1$*$} 
}

\date{
    \textsuperscript{1}Genentech, \textsuperscript{2}Princeton University,
    \textsuperscript{3} MIT,  \\ 
     \textsuperscript{4} Texas A\&M University, 
    \textsuperscript{5} UC Berkeley 
}

\begin{document}

\maketitle

\begin{abstract}
{
This tutorial provides an in-depth guide on inference-time guidance and alignment methods for optimizing downstream reward functions in diffusion models. While diffusion models are renowned for their generative modeling capabilities, practical applications in fields such as biology often require sample generation that maximizes specific metrics (e.g., stability, affinity in proteins, closeness to target structures). In these scenarios, diffusion models can be adapted not only to generate realistic samples but also to explicitly maximize desired measures at inference time without fine-tuning. This tutorial explores the foundational aspects of such inference-time algorithms. We review these methods from a unified perspective, demonstrating that current techniques\textemdash such as Sequential Monte Carlo (SMC)-based guidance, value-based imoprtance sampling, and classifier guidance\textemdash aim to approximate soft optimal denoising processes (a.k.a. policies in RL) that combine pre-trained denoising processes with value functions serving as look-ahead functions that predict from intermediate states to terminal rewards. Within this framework, we present several novel algorithms not yet covered in the literature. Furthermore, we discuss (1) fine-tuning methods combined with inference-time techniques, (2) inference-time algorithms based on search algorithms, which have received limited attention in current research, and (3) connections between inference-time algorithms in language models and diffusion models. The code of this tutorial on protein design is available at \href{https://github.com/masa-ue/AlignInversePro}{https://github.com/masa-ue/AlignInversePro}. 

 } 
\end{abstract}


\newpage 
\section*{Introduction}

Diffusion models \citep{sohl2015deep,ho2020denoising,song2020denoising} have demonstrated remarkable success in computer vision, particularly as generative models for continuous domains such as images \citep{rombach2022high}. This success has been further extended to scientific areas such as the generation of protein 3D structures \citep{yim2023se,watson2023novo,chu2024all,abramson2024accurate} and small molecule 3D structures \citep{xu2022geodiff,jing2022torsional,corso2022diffdock}. Furthermore, 
recent works \citep{shi2024simplified,sahoo2024simple,lou2023discrete} have shown promising results with diffusion over traditional autoregressive models in \emph{discrete} domains. Building on this progress in natural language processing (NLP), the use of diffusion models has also been explored for generating biological sequences (proteins, RNA, and DNA), which are inherently non-causal, because these sequences fold into complex tertiary (3D) structures \citep{campbell2024generative,sarkar2024designing,winnifrith2024generative,wang2024finetuning}.

Controlled generation is a pivotal topic in the study of diffusion models. In the context of ``foundational models'', the process typically begins with training conditional diffusion models on large datasets to generate natural designs (e.g., biologically plausible protein sequences) conditioned on fundamental functionalities. Following this pre-training stage, the focus often shifts to optimizing specific downstream reward functions, commonly referred to as ``alignment'' problems in AI. By guiding generation to maximize a given reward during inference (e.g., binding affinity or stability in protein sequences), diffusion models can be effectively utilized as robust computational design frameworks. Similarly, conditioning on target properties during inference is treated as a reward maximization task, where rewards are frequently defined using classifiers.

\begin{wrapfigure}{r}{0.55\textwidth}
  \begin{center}
    \includegraphics[width=0.54\textwidth]{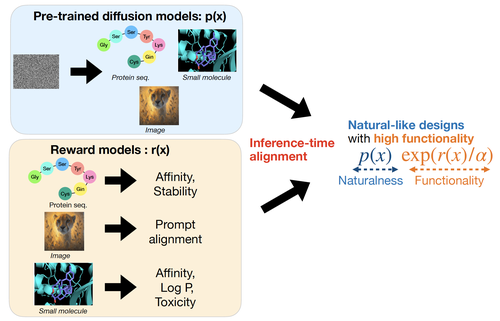}
  \end{center}
  \caption{{The objective of inference-time techniques is to generate natural designs (e.g., natural images or natural-like protein sequences) with high functionality, without any direct fine-tuning of diffusion models. }}
  \label{fig:objective}
\end{wrapfigure}

{ In this tutorial, we aim to explore inference-time techniques for controlled generation in diffusion models, along with their foundational properties. These techniques aim to seamlessly integrate pre-trained generative models trained on large-scale datasets with reward models, as illustrated in \pref{fig:objective}. Specifically, at each generation step in pre-trained diffusion models, certain modifications are introduced to optimize downstream reward functions as summarized in \pref{fig:whoe_summary}. A significant advantage of these methods is that they don't require post-training of the diffusion models, which can be computationally demanding. The simplest such approach is best-of-N sampling in \pref{fig:whoe_summary}a, which involves generating multiple designs (N samples) from a pre-trained diffusion model and selecting the best one based on reward functions (e.g., \citet{nakano2021webgpt}). However, this method can be highly inefficient when the reward functions are difficult to optimize. More efficient sophisticated strategies include classifier guidance in \pref{fig:whoe_summary}b and its variants \citep{dhariwal2021diffusion, song2021score}, sequential Monte Carlo-based methods in \pref{fig:whoe_summary}c \citep{wu2024practical,dou2024diffusion,cardoso2023monte,phillips2024particle}, and value-based sampling methods in \pref{fig:whoe_summary}d
 \citep{li2024derivative}.

\begin{figure}[!t]
\begin{subfigure}{0.5\textwidth}
        \centering
\includegraphics[width=0.78\textwidth]{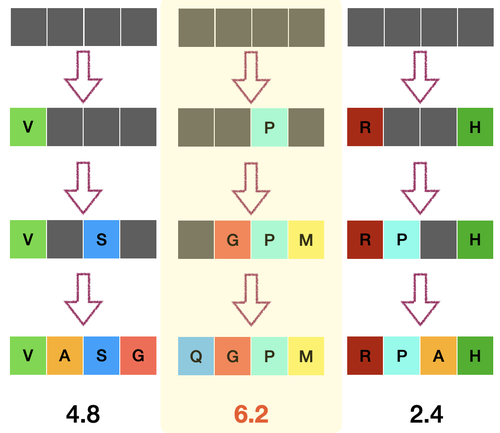}
        \caption{Best-of-N }
    \end{subfigure}%
    \begin{subfigure}{0.5\textwidth}
        \centering
        \includegraphics[width=0.9\textwidth]{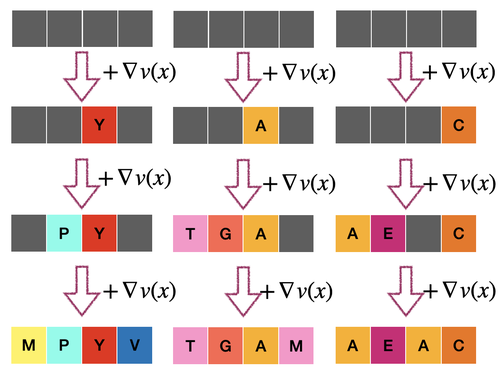}
        \caption{Classfier Guidance (\pref{sec:derivative},\,\ref{sec:derivative_discrete})}
    \end{subfigure}
    \\
    \begin{subfigure}{0.48\textwidth}
        \centering
     \includegraphics[width=0.8\textwidth]{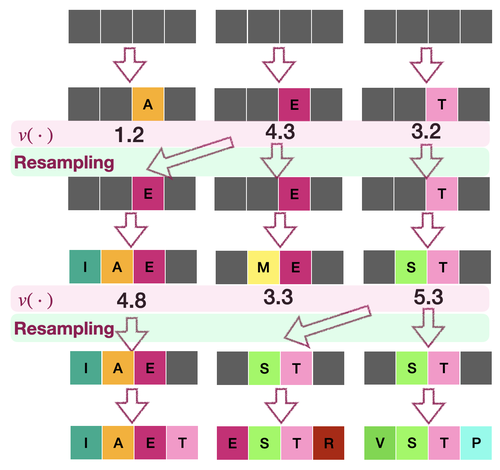}
        \caption{SMC-Based Guidance (\pref{sec:SMC}) }
    \end{subfigure}%
    \begin{subfigure}{0.5\textwidth}
        \centering
        \includegraphics[width=0.9\textwidth]{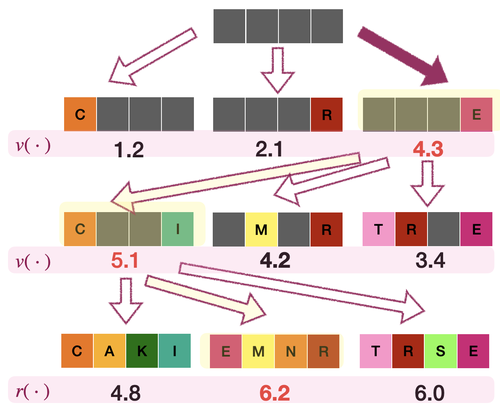}
        \caption{Value-Based Importance Sampling (\pref{sec:SVDD})}
    \end{subfigure}%
\caption{Summary of representative inference-time algorithms. Here,
we aim to optimize downstream reward functions $r:\Xcal \to \RR$ given pre-trained masked diffusion models for sequences. The value function $v(\cdot)$ serves as a look-ahead function, mapping intermediate states to expected future rewards $r(\cdot)$. Best-of-N is a na\"ive method that selects the best sample among $N$ generated ones. Derivative-based guidance adds gradients of differentiable value function models during inference, making it a powerful method when we can construct the actual value function models. SMC-Based Guidance and Value-Based Importance Sampling (a.k.a. beam search with value functions) are gradient-free methods that sequentially select favorable intermediate states based on value functions. These methods do not require constructing differentiable value function models, which can often be challenging in molecular design.
}
\label{fig:whoe_summary}
\end{figure}

Before delving into the details of inference-time techniques, we provide a brief overview of this tutorial in the introduction. We begin by emphasizing the advantages of inference-time methods over post-training approaches, which can also enable controlled generation. Next, we outline the key components essential for inference-time controlled generation. Finally, we offer a comprehensive overview of the inference-time techniques covered in this work.

\begin{figure}[!th]
    \centering
    \begin{subfigure}[b]{1.0\textwidth}
        \centering
         \includegraphics[width=0.7\linewidth]{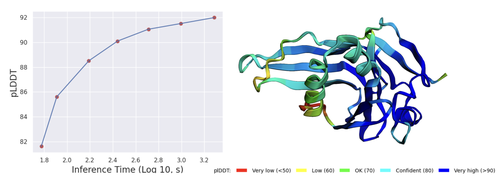}
        \caption{The Y-axis represents the mean pLDDT across residues, which is a common computational proxy for stability derived from AlphaFold2, as noted in \citet{dauparas2022robust,ye2024proteinbench,ahdritz2024openfold} }
        \label{fig:figure1}
    \end{subfigure}
       \begin{subfigure}[b]{1.0\textwidth}
        \centering
         \includegraphics[width=0.65 \linewidth]{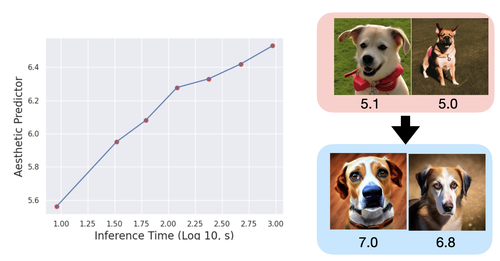}
        \caption{LAION Aesthetic Predictor V2 \citep{schuhmann2022laion} serves as a typical proxy for aesthetic scores. For instance, in the figure on the right, the scores correspond to images in {\color{pink} pink} generated by the pre-trained diffusion model (Stable Diffusion in \citet{podell2023sdxl}), while the scores correspond to images in {\color{columbiablue} blue} obtained using inference-time techniques.}
    \end{subfigure}
     \hfill
    \caption{Scaling inference time compute via value-based beam search \citep{li2024derivative} in \pref{sec:beam}, progressively increasing the tree width (see \pref{fig:whoe_summary}d). These figures demonstrate that as the computational budget (x-axis) increases, rewards (y-axis) can be optimized more effectively. }
    \label{fig:scaling}
\end{figure}

\subsection*{Inference-Time Techniques vs. Post-Training}

After pre-trainig, there are two main approaches for controlled generation: inference-time techniques (i.e., without fine-tuning diffusion models) and post-training methods such as RL-based fine-tuning \citep{black2023training,fan2023dpok,clark2023directly,uehara2024feedback} or classifier-free guidance-based fine-tuning \citep{ho2022classifier,zhang2023adding,yuan2024reward}. In this work, we focus on reviewing the former. For a comprehensive overview of the latter approach, we refer readers to \citet{uehara2024understanding}. While both approaches are important, inference-time techniques generally offer several advantages:
\begin{itemize} 
\item Inference-time techniques are particularly \emph{straightforward to implement}, as many of these methods are not only \emph{fine-tuning-free} but also \emph{training-free}, given access to reward functions. Despite their simplicity, they could deliver competitive performance compared to RL-based fine-tuning approaches. Indeed, for example, we can effectively optimize downstream rewards without any fine-tuning by increasing the computational budget, as illustrated in \pref{fig:scaling}. This scaling approach has recently been further explored in \citet{singhal2025general,ma2025inferencetimescalingdiffusionmodels}.

\item \emph{Inference-time techniques can support post-training}. For example, they can be employed as data augmentation methods within classifier-free guidance or as teacher policies in policy distillation-based post-training. Further details are provided in \pref{sec:compare_distill}.

\item Even after obtaining fine-tuned models through post-training techniques, applying inference-time methods to fine-tune models can be advantageous for further improving the functionality of the generated outputs. This is particularly relevant when downstream reward feedback is highly accurate. Post-training may not fully exploit the information provided by the reward feedback, as it involves converting this feedback into data, a process that can result in information loss. In contrast, inference-time techniques can directly utilize reward feedback without the need for such conversion, enabling more effective optimization.

\end{itemize}

\vspace{-2mm}
\subsection*{Critical Considerations for Choosing Inference-Time Techniques}
\vspace{-2mm}

In this article, we categorize current inference-time techniques according to the following features:

\begin{enumerate}
     \item \textbf{Computational and Memory Efficiency:}  In general, even when utilizing the same algorithm, increased computational or memory resources during inference can result in higher-quality designs. For example, when using beam search with estimated value function, as the beam width increases, the performance increases while the computational time increases as illustrated in \pref{fig:scaling}.   Additionally, the ease of parallel computation is an important practical consideration.
    \item \textbf{What Rewards We Want to Optimize:} In practice, choosing a good reward function that balances accuracy and computational efficiency is crucial. Besides, it is relevant to consider whether the attributes we aim to optimize (referred to as reward models in this draft) function as classifiers, as seen in the standard guidance literature, or as regressors, as is common in the literature on alignment. In this draft, the former task is often called \emph{conditioning}, while the latter is called \emph{alignment}. 
    \item \textbf{Differentiability of Reward Feedback:} In computer vision and NLP, many useful reward feedback is differentiable. However, in scientific domains such as molecular design, much of the useful reward feedback, such as physics-based simulations \citep{salomon2013overview,chaudhury2010pyrosetta,trott2010autodock}), is non-differentiable. Additionally, when utilizing learned reward models as feedback, they are often non-differentiable due to their reliance on non-differentiable features such as molecular fingerprints or biophysical descriptors \citep{stanton1990development,yap2011padel,li2015improving}. This consideration is practically important, as it influences the choice of inference-time technique, as illustrated in \pref{fig:roadmap}.
    
\end{enumerate}

\newpage 
\subsection*{Summary}

\begin{figure}[!th]
    \centering
    \includegraphics[width=0.7\linewidth]{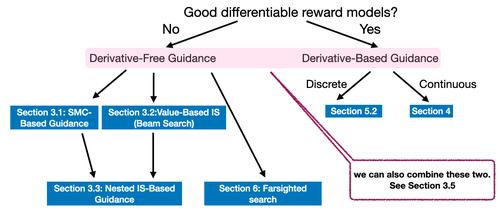}
    \caption{Roadmap of This Paper.}
    \label{fig:roadmap}
\end{figure}

Considering these aspects, we provide a unified categorization of current inference-time techniques in diffusion models, while also highlighting novel perspectives. The key message of this tutorial is summarized as follows. 

\begin{AIbox}{Key Message: A Unified Framework for Inference-Time Techniques in Diffusion Models}
{ All methods introduced here (summarized in \pref{fig:whoe_summary}) generally aim to approximately sample from specific target distributions. Denoting the reward function as $r:\Xcal \to \RR$ (e.g., classifiers or regressors) and $p^{\pre}(\cdot)$ as the distribution induced by policies (i.e., denoising processes) from the pre-trained model, the target distribution is defined as  
\begin{align*}
    \underbrace{p^{\pre}(\cdot)}_{\mathclap{\substack{\textbf{Pre-trained dist. } \\ \text{(Naturalness)}}}} \times \underbrace{\exp(r(\cdot)/\alpha))}_{\mathclap{\substack{\textbf{Reward term } \\ \text{(High functionality)}}}}
    /C\,\,(:= \argmax_{p: \Xcal \to \Delta(\Xcal)}\EE_{x\sim p}[r(x)] -\alpha \mathrm{KL}(p\| p^{\pre})), 
\end{align*}
where $C$ is the normalizing constant, and $\alpha$ is a hyperparameter. This distribution is desirable because the generated outputs exhibit both \textbf{naturalness} and \textbf{high functionality}.

To enable sampling from this distribution, by denoting pre-trained denoising process as $\{p^{\pre}_{t}(\cdot \mid x_{t+1})\}_{t=T}^0$ (from $t=T$ to $t=0$), all methods presented here (methods in \pref{fig:whoe_summary}) employ the following distribution as the denoising process (i.e., optimal policies in RL) at each step during inference:
\begin{align}\label{eq:soft_value}
     p^{\star}_{t-1}(\cdot|x_t):=\underbrace{p^{\pre}_{t-1}(\cdot \mid x_t)}_{\textbf{Pre-trained polices}}\times \underbrace{\exp(v_{t-1}(\cdot)/\alpha)}_{\textbf{Soft value functions}}, 
    \end{align}
where soft value functions act as look-ahead functions that predict the reward at the terminal state $x_0$ from intermediate state $x_t$ (formalized later in \pref{sec:foundation}). The primary distinction among inference-time algorithms lies in \textbf{how this approximation is achieved}, and the effectiveness of each method depends on the specific scenario.  
} 
\end{AIbox}

Additionally, we explore more advanced aspects of inference-time methods in diffusion models, including their integration with fine-tuning, search algorithms, editing, and applications to masked language models beyond diffusion frameworks. The remainder of this tutorial is organized as follows. 
\begin{itemize}
    \item \pref{sec:foundation}: We start by outlining the foundational principles of inference-time techniques. Specifically, we introduce the soft optimal policy defined in \eqref{eq:soft_value}, which represents the denoising process targeted during inference. All methods discussed in this tutorial aim to approximate this optimal policy.
    \item \pref{sec:derivative_free}: We review inference-time techniques that do \emph{not} require differentiable reward feedback, particularly useful in molecular design. These methods are roughly divided into two main categories: the SMC-based approach \citep{wu2024practical,dou2024diffusion,cardoso2023monte,phillips2024particle} and the value-based importance sampling approach \citep{li2024derivative}. 
    Additionally, we explain how to integrate these two approaches. 
    \item \pref{sec:derivative}, \ref{sec:derivative_discrete}: We review methods that require differentiable reward or value function models. This is useful when effective differentiable reward models can be constructed, such as for inpainting tasks in computer vision or motif scaffolding in protein design. We first discuss these methods for diffusion models in the continuous domain in \pref{sec:derivative}, also known as classifier guidance \citep{dhariwal2021diffusion,song2021score}, with their formalization through Doob's transform. Then, we provide an analogous explanation for discrete diffusion models in \pref{sec:derivative_discrete}.
     \item \pref{sec:MCTS}: As mentioned in \pref{sec:derivative_free}, beam search based on value functions is a natural solution for alignment tasks. A next natural step is to integrate more advanced search algorithms. We briefly discuss how search-based algorithms (e.g., MCTS) can be applied to diffusion models. 
      \item \pref{sec:editing}: {The inference-time techniques described in Sections \pref{sec:derivative_free} to \pref{sec:derivative_discrete} primarily focus on generating designs from fully noised states. However, in protein design, objectives often involve editing endogenous designs that meet stringent constraints. We examine how these objectives can be achieved by iteratively adapting the inference-time techniques presented in Sections \pref{sec:derivative_free} to \pref{sec:derivative_discrete} through sequential refinement. } 
    \item  \pref{sec:autoregresstive}: We provide a concise review of inference-time techniques in language models, focusing on autoregressive models like GPT \citep{brown2020language} and masked language models such as BERT \citep{kenton2019bert}, and examine the similarities and differences between autoregressive models, masked language models, and diffusion models. 
     \item \pref{sec:fine_tuning}: We describe how to apply inference-time techniques through distillation to fine-tune diffusion models. This is particularly important given that pure inference-time techniques can result in higher inference costs. We also establish connections between these methods and RL-based fine-tuning approaches. 
    \item \pref{sec:related_works}: We provide a brief overview of relevant algorithmic approaches in protein design, including walk-jump sampling \citep{frey2023protein} and hallucination methods (i.e., genetic algorithms, MCMC-based approaches) \citep{anishchenko2021novo,jendrusch2021alphadesign}.
\end{itemize}

\vspace{-4mm}
{ \section*{Application in Computational Protein Design } 

\begin{wrapfigure}{!r}{0.46\textwidth}
  \begin{center}
    \includegraphics[width=0.44\textwidth]{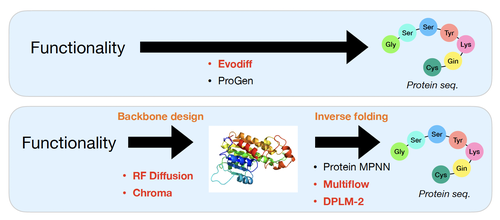}
  \end{center}
    \caption{ Typical foundational diffusion models for protein Sequences (especially, \textcolor{red}{Red} indicates diffusion models). The former approach relies solely on sequence data, while the latter explicitly generates the structure first, followed by the sequence. Notably, hybrid approaches that combine these methods have become increasingly popular in recent studies.  \vspace{-3mm}}
\end{wrapfigure}

While our review primarily focuses on algorithms applicable to general domains, we provide a brief overview of how inference-time techniques can be applied to computational protein design to offer a concrete example. A key step in understanding how inference-time techniques operate is selecting pre-trained diffusion models and defining reward functions, as depicted in \pref{fig:objective}. We discuss several options in the following sections. Code for this tutorial is available at \href{https://github.com/masa-ue/AlignInversePro}{https://github.com/masa-ue/AlignInversePro}.

\paragraph{Pre-Trained Foundational Diffusion Models for Protein Sequences.}

Here, we briefly categorize foundational generative models capable of generating natural-like protein sequences. For additional references, see \citet{winnifrith2024generative}, for example.

1. Purely Sequence-based Models: A common approach involves training discrete diffusion models exclusively on sequence datasets, without using structural datasets, as demonstrated in EvoDiff \citep{alamdari2023protein}. 

2. Structure-based Models: Another widely used approach involves explicit two-step procedures. In the first step, backbone structures are generated using diffusion models such as RFdiffusion \citep{watson2023novo} and Chroma \citep{ingraham2023illuminating}. In the second step, protein sequences are generated conditioned on the backbone structure, a process often referred to as \emph{inverse folding} such as ProteinMPNN \citep{dauparas2022robust}. 

3.  Hybrid Models: Recent approaches leverage structural information in the above but do so seamlessly by jointly generating sequences and structures in each step, as proposed in MultifLow \citep{campbell2024generative}, DPLM-2 \citep{wang2024dplm}, Protein Generator \citep{lisanza2023joint} and ESM3 \citep{hayes2024simulating}.

We note that the above foundational models are often introduced as conditional models. Typical conditions include secondary structures, motifs, domains,  symmetry, etc.

\paragraph{Reward Models (Mapping Protein Sequences to Functionality).}

Here, we briefly outline practical approaches for constructing reward models. Common objectives we aim to optimize include binding affinity, stability, solubility, rigidity, etc. Ideally, reward models should be built using experimental data consisting of $x$ (protein sequences) and $y$ (target properties to be optimized, obtained from experimental assays). However, given the often limited size of such datasets, we frequently leverage features extracted from external ``foundational'' models trained on large datasets or incorporate external knowledge, such as physics-based insights. To construct these features, for example, we can utilize: (1) features derived from representation learning methods, such as masked language models (e.g., ESM \citep{hayes2024simulating}), (2) structural features based on predicted structures, and (3) biophysical features (e.g., $\Delta G$ computed through physics-based simulations).

In the absence of experimental data, it is practical to directly utilize outputs from ``foundational'' models or physics-based models as rewards. For instance, binding affinity can be assessed using outputs from AlphaFold3 \citep{abramson2024accurate} or physics-based software such as Rosetta and AutoDock \citep{chaudhury2010pyrosetta,trott2010autodock}.
}

\tableofcontents

\section{Preliminaries} 

In this section, we begin by discussing the fundamentals of diffusion models. We then formalize our objectives, such as conditioning or alignment (i.e., reward maximization), during inference without fine-tuning. Finally, we introduce several post-training approaches (RL-based fine-tuning and classifier-free guidance) and compare them with purely inference-time techniques.

\subsection{Diffusion Models} 

In diffusion models \citep{sohl2015deep, ho2020denoising, song2020denoising}, the goal is to learn a sampler $p^{\pre}(\cdot) \in \Delta(\mathcal{X})$ for a given design space $\mathcal{X}$ using datasets. For instance, in protein conformation generation, $\mathcal{X}$ may be a Euclidean space if 3D coordinates are used as inputs or a Riemannian manifold if torsion angles are employed, and in protein inverse folding, $\mathcal{X}$ corresponds to a discrete space.

Our objective in training diffusion models is to learn a sequential mapping (i.e., a denoising process) that transitions from a noise distribution to the true data distribution. The training procedure is summarized as follows: Initially, a forward \emph{noising} process $q_t: \mathcal{X} \to \Delta(\mathcal{X})$ is pre-defined, spanning from $t = 0$ to $t = T$. Here, we often refer to such a noising process as a \emph{policy}, following the terminology of RL in our tutorial.

Then, our goal is to learn a reverse \emph{denoising} process ${p_t}$, where each $p_t$ is $\mathcal{X} \to \Delta(\mathcal{X})$, ensuring that the distributions induced by both the forward and backward processes are marginally equivalent. To achieve this, the backward processes are parameterized by $\theta \in \mathbb{R}^d$ through neural networks, and the following loss function is optimized:
\begin{align*} 
 \EE_{x_1,\cdots,x_T\sim q(\cdot|x_0)}\left [-\log p_0(x_0|x_1) + \sum_{t=1}^{T-1}\KL(q_{t}(\cdot \mid x_{t-1},x_0) \| p_{t}(\cdot \mid x_{t+1};\theta)) +  \KL( q_{T}(\cdot) \| p_{T}(\cdot) ) \right], 
\end{align*}
where the expectation is taken with respect to the distribution induced by the forward process. This loss function represents the variational lower bound of the negative log-likelihood $-\log p(x_0)$, commonly known as the ELBO.
  
The remaining question is how to define the noising and denoising processes. Below are examples of concrete parameterizations for these processes in Euclidean space and discrete space.

\begin{Exabox}{XXX}
\begin{example}[Euclidean space] \label{exa:continuous}
When $\mathcal{X}$ is a Euclidean space, we typically use the Gaussian distribution $q_t(\cdot\mid x_{t}) = \Ncal(\sqrt{\alpha_t}x_{t}, (1- \alpha_t)\mathrm{I} )$ as the forward noising process where $I$ is an identity matrix, where $\alpha_t \in \RR$ denote a noise schedule. Then, the backward process is parameterized as 
\begin{align}\label{eq:mean}
    \Ncal\left (\frac{\sqrt{ \alpha_t}(1- \bar \alpha_{t-1}) x_t +  \sqrt{ \bar \alpha_{t-1}}(1-\alpha_t)\hat x_0(x_t;\theta) }{1 - \bar \alpha_t},\sigma^2_t \rI \right),\,\sigma^2_t= \frac{(1-\alpha_t)(1-\bar \alpha_{t-1}) }{1-\bar \alpha_t }, 
\end{align}
where $\bar \alpha_t= \prod_{i=1}^t \alpha_i $. The loss function for $\theta$ is 
\begin{align*}
    \sum_t \EE_{x_t \sim q_t(\cdot|x_0)}[w(t) \|\hat x_0(x_t;\theta)-x_0 \|^2_2]
\end{align*}
where $w$ is a weighting function: $w: [0, T] \to \RR$ and $q_t(\cdot|x_0)$ is a distribution induced by forward policies from $0$ to $t$. { A typical choice of $w(t)$ is $\lambda(t-1)-\lambda(t)$ where $\lambda(t)=\bar \alpha_{t}/(1-\bar \alpha_t)$. 
} 
Here, $\hat{x}_0(x_t; \theta)$ is a neural network that predicts $x_0$ from $x_t$ (i.e., $\EE_{x_0 \sim q(\cdot \mid x_t)}[x_0 \mid x_t]$).
\end{example}
\end{Exabox}

\begin{remark}[Different parameterization]
In the above diffusion models for continuous domains, we also note that there are two additional ways of parameterization. For further details, see \citet{luo2022understanding}. The first parameterization involves predicting the noise $\epsilon_0$ rather than $x_0$, recalling that
$x_t = \sqrt{\bar \alpha_t}x_0 +\sqrt{1-\bar \alpha_t}\epsilon_0, \epsilon_0 \sim \Ncal(0,\rI)$. In this case, using the relation $x_0 =(x_t -\sqrt{1-\bar \alpha_t} \epsilon_0/\sqrt{\bar \alpha_t})$, the mean in the backward denoising process \eqref{eq:mean} is parameterized as 
\begin{align*}
    \frac{1}{\sqrt{\alpha_t}}\left(x_t - \frac{1-\alpha_t}{\sqrt{1-\bar \alpha_t}} \hat \epsilon_0(x_t;\theta)  \right). 
\end{align*}
Another parameterization is the score parameterization, which aims to estimate $\nabla_{x_t} \log q_t(x_t)$ where $q_t$ is a marginal distribution at time $t$ induced by forward noising processes. In this case, the mean in the backward denoising process \eqref{eq:mean} is 
\begin{align*}
      \frac{1}{\sqrt{\alpha_t}}\left(x_t + (1-\alpha_t)\widehat {\nabla_{x_t} \log p}(x_t;\theta)  \right). 
\end{align*}
In particular, when $\alpha_t=1-(\delta t)$ for small $\delta t$, ignoring second-order terms $O((\delta t)^2)$, the above expression simplifies to:
\begin{align}\label{eq:score_rep}
     x_t + (\delta t)\left \{0.5 x_t  + \widehat{\nabla_{x_t} \log p}(x_t;\theta) \right \},\quad \sigma^2_t=(\delta t). 
\end{align}
\end{remark}

\begin{Exabox}{XXX}
\begin{example}[Discrete space (masked diffusion models)]\label{exa:discrete}

Here, we explain masked diffusion models \citep{sahoo2024simple, shi2024simplified}, inspired by seminal works on discrete diffusion models (e.g., \citet{ austin2021structured,campbell2022continuous,lou2023discrete}).

Let $\Xcal$ be a space of one-hot column vectors $\{x\in\{0,1\}^K:\sum_{i=1}^K x_i =1\}$, and $\mathrm{Cat}(\pi)$ be the categorical distribution over $K$ classes with probabilities given by $\pi \in \Delta^K$ where $\Delta^K$ denotes the K-simplex. A typical choice of the forward noising process is  
$q(x_{t+1}\mid x_{t}) = \mathrm{Cat}(\alpha_t x_{t}+ (1- \alpha_t)\mathbf{m} )$
where $\mathbf{m}=[\underbrace{0,\cdots,0}_{K-1},\mathrm{Mask}]$. Then, the backward process is parameterized as
\begin{align*}
x_{t-1}=
   \begin{cases}  
     \delta(\cdot = x_t) \quad \mathrm{if}\, x_t\neq \mathbf{m}  \\
       \mathrm{Cat}\left ( \frac{(1-\bar \alpha_{t-1})\mathbf{m}  + (\bar \alpha_{t-1}- \bar \alpha_t) \hat x_0(x_t;\theta) }{ 1 - \bar \alpha_t } \right),\quad \mathrm{if}\,x_t= \mathbf{m}, 
   \end{cases}  
\end{align*}
where $\bar{\alpha}_t = \Pi_{i=1}^t \alpha_i$. The loss function for $\theta$ is 
\begin{align}\label{eq:loss_discrete}
    \sum_t \EE_{x_t \sim q_t(\cdot|x_0)}[w(t)\mathrm{I}(x_t=\mathbf{m})\langle x_0, \log \hat x_0(x_t;\theta)\rangle  ] 
\end{align}
where $w$ is a weight function. A typical choice of $w(t)$ is $\lambda(t-1)-\lambda(t)$ where $\lambda(t)=\bar \alpha_{t}/(1-\bar \alpha_t)$.

Here, $\hat{x}_0(x_t; \theta)$ is a neural network that predicts $x_0$ from $x_t$ (when $x_t$ is masked). Note that the above loss function is similar to the one used in BERT \citep{devlin2018bert}. Compared to BERT, masked diffusion models are more hierarchical, with a greater variety of masking modes. Given the success of BERT-type algorithms for biological sequences, such as ESM \citep{hayes2024simulating} for protein, DNA BERT \citep{ji2021dnabert}, RNA BERT \citep{akiyama2022informative}, CodonBERT \citep{ren2024codonbert}, the application of discrete diffusion models to biological sequences is considered as a natural extension.

\end{example}
\end{Exabox}

\begin{remark}{Multiple tokens} In the above, we focused on the case of a single token. When considering a sequence of $L$ tokens ($x^{1:L})$, we employ noising/denoising processes defined as $p_{t-1}(x^{1:L}_{t-1} |x^{1:L}_{t}) = \prod_{l=1}^L p_{t-1}(x^l_{t-1}|x^{1:L}_{t})$, which is  independent across tokens. This independence is crucial for mitigating the curse of token length, i.e., avoiding the need to directly model the entire space with a cardinality of $O(K^L)$. This aspect also plays a significant role in the inference-time techniques discussed later in \pref{sec:discrete_first}, as also highlighted in \citet{nisonoff2024unlocking}. 
\end{remark}

After learning the backward process as in the above examples, we can sample from a distribution that emulates training data distribution (\textit{i.e.}, $p^{\pre}(x)$) by sequentially sampling $\{p_t\}_{t=T}^0$ from $t=T$ to $t=0$. In this draft, given a pre-trained model $\{p^{\pre}_t\}_{t=T}^0$, we denote the induced distribution by $p^{\pre}(x)$, i.e., $$p^{\pre}(x_0)=\int \left \{ \prod_{t=T+1}^1 p^{\pre}_{t-1}(x_{t-1}|x_t)\right \}dx_{1:T}. $$
Here, with slight abuse of notation, we denote $p^{\pre}_{T}(\cdot \mid \cdot)$ by $p^{\pre}_T(\cdot)$. 

\paragraph{Conditional Generative Models.}
Note that while the following discussion assumes pre-trained models are unconditional generative models, it can be easily extended to cases where pre-trained models are conditional generative models, $p(x \mid c): \mathcal{C} \to \Delta(\mathcal{X})$. For instance, in protein conformation generation, $c$ represents a 1D amino acid sequence, and $x$ is the corresponding protein conformation. In protein inverse folding, $c$ typically refers to a backbone structure, and $x$ is a 1D amino acid sequence. The training process for conditional diffusion models is nearly identical to that of unconditional models, with the only difference being the need to augment the input of the denoising process with the additional space $\mathcal{C}$. In this process, we typically introduce an additional ``unconditional class'', as implemented in classifier-free guidance \citep{ho2020denoising}.

\paragraph{Notation.} The notation $\delta_{a}$ denotes a Dirac delta distribution centered at $a$. The notation $\propto$ indicates that the distribution is equal up to a normalizing constant. With slight abuse of notation, we often denote $p_T(\cdot|\cdot)$ by $p_{T}(\cdot)$. The notation $\mathrm{Cat}([w_1,\cdots,w_M])$ means a categorical distribution with probability $[w_1,\cdots,w_M]$.

\subsection{Objectives: Conditioning, Inverse Problems, Reward Maximization }
\label{sec:goal}

In this review, we consider a scenario where a pre-trained diffusion model has been trained on a large dataset. For example, this could be an unconditional generative model that captures the natural-like protein space. Then, at inference time (during sampling), we aim to guide the generation process to achieve a specific objective. Generally, we seek to optimize a reward function $r: \mathcal{X} \to \mathbb{R}$ that characterizes this objective, such as the binding affinity of proteins. Later, we will discuss how to define such rewards in more detail.  Mathematically, the goal above is formalized as follows. 

\begin{AIbox}{Overall General Objectives}
We aim to generate designs that achieve high rewards while preserving the naturalness of the samples. More specifically, our goal is to sample from the distribution:
\begin{align}
 p^{(\alpha)}(\cdot)& := \frac{\exp(r(\cdot )/\alpha)p^{\pre}(\cdot)}{\int \exp(r(x)/\alpha)p^{\pre}(x)dx}=  \argmax_{p\in [\Delta(\Xcal)] }\underbrace{\EE_{x\sim p(\cdot)}[r(x)]}_{\textbf{Term (a): Reward}}-\alpha \underbrace{\KL( p(\cdot) \| p^{\pre}(\cdot))}_{\textbf{Term(b): Naturalness}}  \label{eq:target_dist}. 
\end{align}
Term (a) is introduced to optimize the reward function, while term (b) ensures the naturalness of the generated samples. For instance, in protein design, this formulation aligns with the objective of generating natural-like proteins that exhibit high functionality. 
\end{AIbox}

{ In the following, we provide detailed examples of reward functions, categorized into three primary scenarios: (a) conditioning problems, (b) inverse problems, and (c) alignment problems. To summarize, in conditioning, rewards are typically classifiers; in inverse problems, rewards are (known) likelihoods; and in alignment problems, rewards are regressors. However, it is important to note that these scenarios are often treated in a mixed manner in the literature.  }

\subsubsection{Conditioning}\label{sec:conditioning}

In scenarios where we aim to sample from conditional distributions $p(x \mid y) \in [\mathcal{Y} \to \Delta(\mathcal{X})]$, a natural choice for the reward function $r$ is the log-likelihood $\log p(y \mid x)$, defined by a classifier $p(y \mid x)$. This classifier is often not predefined and must be learned from datasets consisting of paired samples $(x, y)$. For example, in protein design, classifiers that categorize protein functions/families (such as enzymes, transport proteins, and receptors) serve as effective reward functions. In particular, setting $\alpha = 1$ enables the generation of samples directly from $p(x \mid y)$, as demonstrated by the following derivation:
\begin{align*}
    p^{(1)}(x|y) \propto \exp(\log p(y|x))p^{\pre}(x)=p(y|x)p^{\pre}(x) \underbrace{\propto}_{\mathrm{Bayes\,theorem}} p(x|y). 
\end{align*}

\subsubsection{Inverse Problems}

A closely related challenge to the conditioning problem is the inverse problem, where the objective is to sample from $p(x \mid y) \in [\mathcal{Y} \to \Delta(\mathcal{X})]$. In inverse problems, we typically assume a known observation model:
\begin{align*} 
y = \Acal(x) + n, 
\end{align*}
where $\Acal: \mathcal{X} \to \mathcal{Y}$ represents the measurement operator, $n$ denotes measurement noise, and $\mathcal{Y}$ is a possibly high-dimensional continuous space. When $n$ follows a Gaussian distribution, the log-likelihood can be expressed explicitly, resulting in the following reward function: $\|y - \Acal(x)\|_2^2$. 

This type of problem frequently arises in computer vision applications, where constructing a conditional model $p(x \mid y)$ without classifier-free guidance is particularly challenging due to the high dimensionality of $y$. A canonical example of an inverse problem is image inpainting, where the objective is to restore missing or corrupted regions of an image in a visually coherent manner. In the field of protein design, analogous tasks include motif scaffolding, where the goal is to generate a scaffold (the complete stable backbone structure) conditioned on the motif (functionally critical structural or sequence elements).

\subsubsection{Reward Maximization (i.e., Alignment)}\label{sec:alignment}

The primary objective in alignment problems is to optimize specific downstream reward functions, often represented by regressors. This task is critical in natural language processing (NLP) and computer vision (CV), commonly referred to as alignment. In protein design, it typically involves maximizing metrics such as binding affinity, immunogenicity, designability, self-consistency, and stability \citep{hayes2024simulating,wang2024finetuning,widatalla2024aligning,ingraham2023illuminating,hie2022high}.

Below, we outline several key considerations:
\begin{itemize} \item \textbf{Known or Unknown Rewards:} In many cases, reward functions can be predefined (i.e., not requiring learning) when optimizing metrics such as docking scores and stability from physics-based simulations or properties like SA, QED, and logP calculated using RDKit in molecular design. However, in many other scenarios, these rewards must be learned from experimental data.
\item \textbf{Quality vs. Computational Efficiency:} 
Even when rewards are directly obtained from physics-based simulations, high-resolution simulations can be computationally prohibitive. For instance, Glide \citep{halgren2004glide} offers high accuracy, while AutoDock Vina \citep{trott2010autodock} is significantly faster but sacrifices accuracy. When selecting specific objectives, balancing the trade-off between computational efficiency and accuracy is a crucial practical challenge.

\item \textbf{Multi-Objectiveness:} 
In molecular design, it is often necessary to optimize multiple objective functions simultaneously. For example, in small molecule generation, the goal is to control properties such as affinity, specificity, designability, lipophilicity, and toxicity. However, since these reward functions often conflict, achieving optimal trade-offs poses a significant challenge.

\end{itemize}

{ \section{Foundations of Inference-Time Controlled Generation}\label{sec:foundation}
} 

In this section, we establish the foundation of inference-time techniques in diffusion models. Specifically, we describe the form of the optimal policy (i.e., denoising process) for the objective introduced in Section~\ref{sec:goal}, which involves guiding natural-like designs with high functionality. All the methods presented in the following sections, from \pref{sec:derivative_free} to \pref{sec:derivative_discrete}, aim to approximate this policy.

\subsection{Soft Optimal Policies (Denoising Processes) in Diffusion Models}\label{sec:soft}

We introduce several key concepts essential for understanding our target policies (i.e., target denoising processes) for inference-time techniques that are detailed in the following section. These concepts are well established through the lens of RL, recognizing that diffusion models can be framed with MDPs. For further details, refer to \citet[Section 3]{uehara2024understanding}. A summary is provided below.

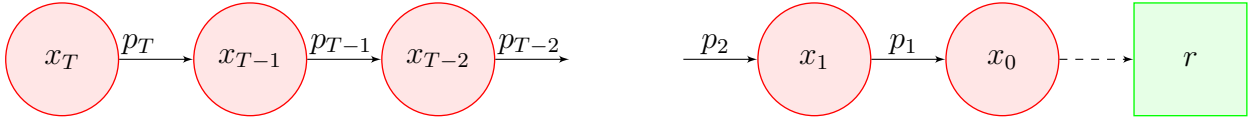
\begin{figure}[!th]
\centering
\begin{tikzpicture}[%
>=latex',node distance=2.5cm, minimum height=1.5cm, minimum width=1.5cm,
state/.style={draw, shape=circle, draw=red, fill=red!10, line width=0.5pt},
state-emp/.style={draw, shape=rectangle, draw=red, fill=red!10, line width=0.5pt,opacity=0.0},
reward/.style={draw, shape=rectangle, draw=green, fill=green!10, line width=0.5pt}
]
\node[state] (xT) at (0,0) {$x_{T}$};
\node[state,right of=xT] (xT1) {$x_{T-1}$};
\node[state,right of=xT1] (xT2) {$x_{T-2}$};
\node[state-emp,right of=xT2] (xT3) {...};
\node[state] (x1) at (10,0) {$x_{1}$};
\node[state,right of=x1] (x0) {$x_{0}$};
\node[reward,right of=x0] (r0) {$r$};
\draw[->] (xT) -- (xT1);
\node[text width=2cm] at (1.8,0.2) {$p_T$};
\draw[->] (xT1) -- (xT2);
\node[text width=2cm] at (4.3,0.2) {$p_{T-1}$};
\draw[->] (xT2) -- (xT3);
\node[text width=2cm] at (6.8,0.2) {$p_{T-2}$};
\draw[->] (xT3) -- (x1);
\node[text width=2cm] at (9.5,0.2) {$p_{2}$};
\draw[->] (x1) -- (x0);
\node[text width=2cm] at (12.0,0.2) {$p_{1}$};
\draw[dashed,->] (x0) -- (r0);
\end{tikzpicture}
\caption{Formulating diffusion models using MDPs.}
\end{figure}

\paragraph{Soft Value Function.} For $t \in [T+1, \cdots, 1]$, we define the \emph{soft value function} as:
\begin{align}\label{eq:soft}
     v_{t-1}(\cdot ):=\alpha \log \EE_{x_0 \sim p^{\pre}(x_0|x_{t-1})}\left [\exp\left (\frac{r(x_0)}{\alpha} \right)|x_{t-1}=\cdot \right ], 
\end{align}
where $\EE_{x_0\sim p^{\pre}(x_0|x_{t-1})\}}[\cdot]$ is induced by pre-trained polices $\{ p^{\pre}_{t}(\cdot|x_{t+1})\}_{t=T}^0$. Since this plays a key role in our tutorial, we highlight it as follows: 

\begin{AIbox}{Soft Value Function} This soft value function captures the expected future reward at $t=0$ from the intermediate state at $t-1$. Intuitively, it serves as a look-ahead function that predicts future rewards $r(x_0)$ based on the intermediate state $x_{t-1}$. It enables us to guide an inference procedure efficiently from $t=0$ to $t=T$, as we will see shortly.  
\end{AIbox}

To further illustrate, consider a scenario where $r$ is a classifier $r(\cdot )=\log p(y|\cdot)$ for class $y$ and $\alpha = 1$. In this case, the above expression simplifies to:
\begin{align*}
    v_{t-1}(\cdot) = \log \EE_{x_0 \sim p^{\pre}(x_0|x_{t-1})}[p(y|x_0)|x_{t-1}=\cdot ], 
\end{align*}
which aligns with the formulation used in classifier guidance \citep{dhariwal2021diffusion,song2021score}.

\paragraph{Soft-Bellman Equation.} Soft-Bellman equations (e.g., \citet{haarnoja2017reinforcement,geist2019theory}) characterize value functions recursively as follows:
\begin{align}\label{eq:soft_bellman}
    \int p^{\pre}_{t-1}(x|x_{t})\exp(v_{t-1}(x)/\alpha)dx = \exp(v_{t}(x_t)/\alpha). 
\end{align}
By iterating the above equation, we can easily derive \eqref{eq:soft} as follows:
\begin{align*}
    \exp\left ( \frac{v_{t}(x_t)}{\alpha} \right) &= \EE_{p^{\pre}}\left [ \exp\left (\frac{v_{t-1}(x_{t-1})}{\alpha} \right)|x_t \right ]=\EE_{p^{\pre}}\left [\exp\left ( \frac{v_{t-2}(x_{t-2})}{\alpha } \right) |x_t\right ]=\cdots \\
     &=\EE_{p^{\pre}}\left [ \exp\left (\frac{r(x_{0})}{\alpha}\right ) |x_t \right ]. 
\end{align*}

\paragraph{Soft Optimal Policy.} We define the \emph{soft optimal policy}  $p^{\star}_{t-1}:\Xcal  \to \Delta(\Xcal)$ as a weighted pre-trained policy with value functions $v_{t-1}:\Xcal \to \RR$: 
\begin{align}\label{eq:optimal_policy}
p^{\star}_{t-1}(\cdot|x_{t})&=\frac{p^{\pre}_{t-1}(\cdot |x_{t})\exp(v_{t-1}(\cdot)/\alpha)}{\int_{x\in \Xcal} p^{\pre}_{t-1}(x|x_{t})\exp(v_{t-1}(x)/\alpha)dx }. 
\end{align}
Intuitively, this policy retains the characteristics of the pre-trained policy while guiding generation toward optimizing reward functions through soft value functions. Using the soft-Bellman equation \eqref{eq:soft_bellman} and  substituting  the denominator in \eqref{eq:optimal_policy}, the above optimal policy simplifies to 
\begin{align*}
     p^{\star}_{t-1}(\cdot|x_{t})=\frac{p^{\pre}_{t-1}(\cdot |x_{t})\exp(v_{t-1}(\cdot)/\alpha)}{\exp(v_{t}(x_t)/\alpha) }.  
\end{align*}
The term ``optimal'' reflects that this policy maximizes the following entropy-regularized objective: 
\begin{align*}
   \argmax_{\{p_t:\Xcal \to \Delta(\Xcal) \}_{t=T}^0 } \EE_{ \{p_t\}_{t=T}^0  } \left[r(x_0)- \alpha \sum_{t=T}^0 \KL(p_{t-1}(\cdot|x_t) \| p^{\pre}_{t-1}(\cdot|x_t))   \right], 
\end{align*}
which is a standard objective in RL-based fine-tuning for diffusion models \citep{fan2023dpok,uehara2024understanding}.

With this setup, we present the following key theorem:

\begin{Thmbox}{}
\begin{theorem}[From Theorem 1 in \citet{uehara2024bridging}]\label{thm:key}
The distribution induced by $\{ p^{\star}_t(\cdot|x_{t+1})\}_{t=T}^0$ (i.e., $\int \left\{ \prod_{t=T+1}^1 p^{\star}_{t-1}(x_{t-1}|x_t)\right \}d x_{1:T}$) is the target distribution $p^{(\alpha)}(\cdot)$ in \eqref{eq:target_dist}, i.e., 
\begin{align*} 
\frac{\exp(r(\cdot )/\alpha)p^{\pre}(\cdot)}{\int_{x\in \Xcal} \exp(r(x)/\alpha)p^{\pre}(x)dx}. 
\end{align*}
\end{theorem}
\end{Thmbox}

Thus, by sequentially sampling from soft optimal policies during inference, we can generate natural-like designs with high functionality. However, this approach presents practical challenges due to two key factors:
\begin{enumerate}
    \item The absence of an exact value function $v_t(\cdot)$
    \item  { The large action space $\Xcal$, making it challenging to compute the normalizing constant in the denominator (i.e., $\int_{x\in \Xcal} \exp(r(x)/\alpha)p^{\pre}(x)dx$) or evaluate value functions in the numerator within a reasonable computational time. } 
\end{enumerate}
In the following section, we first review methods that address the first challenge of value function approximation in \pref{sec:method_value}, followed by approaches that attempt to address the second challenge using these approximated value functions in \pref{sec:derivative_free}-\pref{sec:derivative_discrete}. The key message is as follows. 

\begin{AIbox}{Takeaways}
Each guidance method discussed in this tutorial aims to approximate the soft-optimal policy through different approaches. The method in \pref{sec:derivative_free} seeks to achieve this approximation in a derivative-free manner, while the methods in \pref{sec:derivative} and \ref{sec:derivative_discrete} leverage the gradient of value functions to approximate the policy.
\end{AIbox}

Before addressing solutions to the first challenge, we highlight several additional points related to \pref{thm:key}. 
\begin{itemize} 
 \item \textbf{Extension from Discrete-Time to Continuous-Time Formulation:} Although \pref{thm:key} is formulated for discrete-time diffusion models, it has been extended to continuous-time diffusion models in later work. See \citet[Theorem 1]{uehara2024finetuning} for continuous space diffusion models and \citet[Theorem 1]{wang2024finetuning} for discrete space diffusion models.
\item \textbf{RL-Based Fine-Tuning:} In RL-based fine-tuning, this framework is implicitly employed as the target policy. However, practical implementations differ due to various associated errors (e.g., optimization or function approximation errors), and value functions are typically not directly employed in RL-based fine-tuning.
\item \textbf{Avoiding Curse of Token Length in Discrete Diffusion Models:} In discrete diffusion models (Example~\ref{exa:discrete}), the space $\Xcal$ appears to be $K^L$, where $K$ is the vocabulary size and $L$ is the token length. However, the effective action space is actually $KL$, enabling exact sampling from optimal policies (with approximated value functions) when $L$ and $K$ are relatively small. We will discuss this aspect in more detail in \pref{sec:discrete_first}. 
\item \textbf{For Pre-Defined MDPs:} Our results can be readily extended to scenarios where MDPs are pre-defined, particularly in synthesizable molecular generation \citep{cretu2024synflownet,seo2024generative}. This setting has been widely explored in the literature on GFlowNets \citep{bengio2023gflownet}.
\item \textbf{Relation with Other Works:} Similar (and essentially equivalent) results to \pref{thm:key} have been known in various fields, including soft RL \citep{haarnoja2017reinforcement,levine2018reinforcement}, language models \citep{mudgal2023controlled,han2024value}. In addition, in the computational statistics literature, soft value functions are referred to as twisting potentials, while soft optimal policies are known as optimally twisted policies \citep{doucet2009tutorial,naesseth2019elements,heng2020controlled}.  
\end{itemize}

\subsection{Approximating Soft Value Functions}\label{sec:method_value}

We present several representative methods for approximating value functions, each of which can be integrated into inference-time techniques. They are used for the inference-time technique discussed in subsequent sections, recalling that the inference-time technique aims to approximate soft-optimal policies, which are expressed as the product of value functions and pre-trained policies.

\subsubsection{Posterior Mean Approximation}\label{sec:posterior_mean}

The most straightforward approach is to use the following approximation:
\begin{align*}
    \EE_{x_0\sim p^{\pre}(x_t) } [r(x_0)|x_t]=\int r(x_0)p^{\pre}(x_0|x_t)dx_0\approx r(\EE_{x_0\sim p^{\pre}(x_t) }[x_0|x_t]). 
\end{align*}
In pre-trained models, we have a decoder from $x_t$ to $x_0$, denoted by $\hat x_0(x_t)$ in Example~\ref{exa:continuous} and \ref{exa:discrete}. Thus, a natural approach is to use $r(\hat{x}_0(x_t))$. The whole algorithm is summarized in \pref{alg:PM}. This approximation has been applied for conditioning where the reward functions are classifiers, such as in DPS \citep{chung2022diffusion}, universal guidance \citep{bansal2023universal}, and reconstruction guidance \citep{ho2022video}. 

\begin{algorithm}[!th]
\caption{Value Function Estimation using Posterior Mean Approximation}\label{alg:PM}

\begin{algorithmic}[!th] 
     \STATE {\bf Require}:  Pre-trained diffusion models, reward $r:\Xcal \to \RR$
    \STATE  Set $\hat v^{\diamond}(\cdot,t):= r(\hat x_0(x_t=\cdot))$ 
      \STATE {\bf Output}: $ \hat v^{\diamond}$
\end{algorithmic}
\end{algorithm}

\subsubsection{Monte Carlo Regression}

Another natural approach is to train a value function regressor. From the definition \eqref{eq:soft} of soft value functions, we can show:
\begin{align*}
    \exp(v(\cdot)/\alpha)=\argmin_{f: \Xcal \to \RR}\EE_{x_0\sim p^{\pre}(x_t), x_t\sim u_t}[\{\exp(r(x_0)/\alpha)- f(x_t)\}^2],   
\end{align*}
where $u_t \in \Delta(\Xcal)$ represents a roll-in distribution. By replacing this approximation with an empirical objective and applying function approximation, the complete algorithm is summarized in \pref{alg:MC} when we use distributions induced by pre-trained models as roll-in distributions.

\begin{algorithm}[!th]
\caption{Value Function Estimation using Monte Carlo Regression}\label{alg:MC}
\begin{algorithmic}[1]
     \STATE {\bf Require}:  Pre-trained diffusion models $\{p^{\pre}_t\}^0_{t=T}$, reward $r:\Xcal \to \RR$, function class $\Phi: \Xcal  \times [0,T]\to \RR$.
    \STATE Collect datasets $\{x^{(s)}_{T},\cdots,x^{(s)}_0\}_{s=1}^S$ by rolling-out $\{ p^{\pre}_t\}_{t=T}^0$ from $t=T$ to $t=0$. 
    \STATE 
\begin{align*}
     \hat f = \argmin_{f \in \Phi}\sum_{t=T}^{0} \sum_{s=1}^S \left \{\exp\left (\frac{r(x^{(s)}_0)}{\alpha} \right) - \exp \left ( \frac{f(x^{(s)}_t, t)}{\alpha}\right) \right \}^2. 
\end{align*}
      \STATE {\bf Output}: $\hat f(\cdot)$
\end{algorithmic}
\end{algorithm}

\subsubsection{Soft Q-Learning}

Another natural approach is to use soft Q-learning \citep{haarnoja2017reinforcement,levine2018reinforcement}. Recall the soft-Bellman equations \eqref{eq:soft_bellman}, we have:
\begin{align*}
    \exp\left (\frac{v_t(\cdot)}{\alpha} \right ) =\argmin_{f:\Xcal \to \RR}\EE_{x_t \sim u_t} \left [\left \{\exp\left (\frac{v_{t-1}(x_{t-1})}{\alpha} \right )-f(x_t) \right \}^2 \right], 
\end{align*}
where $u_t \in \Delta(\Xcal)$ is a roll-in distribution. Although the right-hand side is not directly accessible, we can estimate the value functions by recursively applying regression, a procedure known as Fitted Q-iteration (FQI) in RL \citep{ernst2005tree,mnih2015human}. The complete algorithm is summarized in \pref{alg:FQI}.

\begin{algorithm}[!ht]
\caption{Value Function Estimation using Soft Q-learning}\label{alg:FQI}
\begin{algorithmic}[1]
     \STATE {\bf Require}:  Pre-trained diffusion models $\{p^{\pre}_t\}^0_{t=T}$, value function class $\Phi:\Xcal \times [0,T] \to \RR$
    \STATE Collect datasets $\{x^{(s)}_{T},\cdots,x^{(s)}_0\}_{s=1}^S$ by rolling-out $\{ p^{\pre}_t\}_{t=T}^0$ from $t=T$ to $t=0$. 
      \FOR{$j \in [0,\cdots, J] $}
    \STATE  Run regression:  
    \begin{align*}
          \hat f^{j} \leftarrow \argmin_{f \in \Phi}\sum_{t=T}^{0} \sum_{s=1}^S \left \{ \exp\left( \frac{f (x^{(s)}_t)}{\alpha} \right ) - \exp\left (\frac{ \hat f^{j-1}(x^{(s)}_{t-1})}{\alpha}\right ) \right \}^2.
    \end{align*}
     \ENDFOR  
      \STATE {\bf Output}:  $\alpha \log \hat f^{J}$ 
\end{algorithmic}
\end{algorithm}

\section{Derivative-Free Guidance}\label{sec:derivative_free}

We begin by outlining two primary derivative-free approaches that do not rely on differentiable models. As discussed in the introduction, constructing differentiable models in molecular design can be challenging due to the non-differentiable nature of reward feedback for the following reasons. 

\begin{AIbox}{
Scenarios where Non-Differentiable Reward Feedback is Beneficial}
\begin{itemize}
    \item Reward feedback is often provided through black-box physics-based simulations (e.g., docking simulations). 
    \item Non-differentiable features (e.g., molecular descriptors) may be required to build reward models.
    \item Reward models may involve non-differentiable architectures, such as specified graph neural networks or XGBoost.
\end{itemize}    
\end{AIbox}

Therefore, these derivative-free methods are particularly useful in such scenarios. In this section, we first introduce sequential Monte Carlo (SMC)-based guidance, followed by value-based importance sampling approach.

\subsection{SMC-Based Guidance}\label{sec:SMC}

We first provide an intuitive explanation of SMC-based guidance proposed in \citet{wu2024practical,dou2024diffusion,cardoso2023monte,phillips2024particle}, which combine SMC (a.k.a. particle filter) \citep{gordon1993novel,kitagawa1993monte,del2006sequential} with diffusion models. While there are several variants, we focus here on the simplest version. Since SMC-based guidance is an iterative method,  let us consider the process at $t$.  At this stage, we assume there are $N$ samples (particles), $\{x^{[i]}_{t}\}_{i=1}^N$, each with uniform weights: $
1/N\sum_{i=1}^N \delta_{x^{[i]}_{t}}. $
Given this distribution, our goal is to sample from the optimal policy $p^{\star}_{t-1}(\cdot \mid \cdot)$.

For this purpose, using a proposal distribution $q_{t-1}(\cdot | x^{[i]}_{t}):\Xcal \to \Delta(\Xcal)$, such as policies from pre-trained models (we will discuss this choice further in \pref{sec:choice}), we generate new samples $\{\bar x^{[i]}_{t-1}\}$. Ideally, we aim to sample from the optimal policy in \eqref{eq:soft}. To approximate this, based on importance sampling, we consider the following weighted empirical distribution:
\begin{align*}
     \sum_{i=1}^N \frac{w^{[i]}_{t-1}  }{\sum_{j=1}^N w^{[j]}_{t-1}}\delta_{ \bar x^{[i]}_{t-1}}, \quad w^{[i]}_{t-1} = \frac{p^{\star}_{t-1}(\bar x^{[i]}_{t-1}|x^{[i]}_{t} )  }{q_{t-1}( \bar x^{[i]}_{t-1}|x^{[i]}_{t} ) } = \frac{ p^{\pre}_{t-1}(\bar x^{[i]}_{t-1}|x^{[i]}_{t} )\exp(v(\bar x^{[i]}_{t-1})/\alpha) }{q_{t-1}(\bar x^{[i]}_{t-1}|x^{[i]}_{t} ) \exp(v(x^{[i]}_{t})/\alpha )   }. 
\end{align*}
However, as the weights become increasingly non-uniform, the approximation quality deteriorates. To mitigate this, SMC performs resampling with replacement, generating an equally weighted Dirac delta distribution: 
\begin{align}\label{eq:resampling}
    \frac{1}{N} \sum_{i=1}^N \delta_{x^{[i]}_{t-1}}, x^{[i]}_{t-1}:= \bar x^{[\zeta_i]}_{t-1}, \zeta_i \sim \mathrm{Cat}\left [\left\{   \frac{w^{[i]}_{t-1}  }{\sum_{j=1}^N w^{[j]}_{t-1}} \right\}_{i=1}^N \right ]. 
\end{align}

The complete algorithm is summarized in \pref{alg:SMC}. Note that to maintain computational efficiency, the resampling step \eqref{eq:resampling} is executed when the effective sample size, which indicates the degree of weight uniformity, falls below a predefined threshold.

\begin{algorithm}[!th]
\caption{SMC-Based Guidance (e.g.,  \citet{wu2024practical,dou2024diffusion,cardoso2023monte,phillips2024particle})}\label{alg:SMC}
\begin{algorithmic}[1]
     \STATE {\bf Require}: Estimated (potentially \textbf{non-differentiable}) value functions $\{\hat v_t(x)\}_{t=T}^0$ (\pref{sec:method_value}), pre-trained diffusion models $\{p^{\pre}_t\}_{t=T}^0$, proposal polices $\{q_t\}_{t=T}^0$, hyperparameter $\alpha \in \RR$, Batch size $N$
     \FOR{$t\in [T+1,\cdots, 1]$}
       \STATE \textbf{Importance sampling step:}  \label{line:IS} 
Generate $i \in [1,\cdots, N]; x^{[i]}_{t-1} \sim q_{t-1}(\cdot| x^{[i]}_{t})$
       \STATE Update weights:
       \begin{align*}
          w^{[i]} \leftarrow  \frac{ p^{\pre}_{t-1}(x^{[i]}_{t-1}|x^{[i]}_{t} )\exp(\hat v_{t-1}(x^{[i]}_{t-1})/\alpha) }{q_{t-1}(x^{[i]}_{t-1}|x^{[i]}_{t} )\exp(\hat v_t(x^{[i]}_{t})/\alpha)  }w^{[i]}.  
       \end{align*} 
        \STATE \textbf{Resampling step:} Calculate the effective sample size $
            \frac{\{\sum_i w^{[i]} \} ^2  }{\sum \{ w^{[i]}\}^2}.$
        If it falls below a certain threshold, resample by selecting new indices with replacement:  \\ 
       $$\{x^{[i]}_{t-1}\}_{i=1}^N  \leftarrow \{x^{\zeta^{[i]}_{t-1}}_{t-1}\}_{i=1}^N,\quad \{\zeta^{[i]}_{t-1}\}_{i=1}^N \sim \mathrm{Cat}\left ( \left \{\frac{w^{[i]}_{t-1}}{\sum_{j=1}^N w^{[j]}_{t-1} } \right \}_{i=1}^N\right).$$\label{line:selection}
     \ENDFOR
  \STATE {\bf Output}: $\{ x^{[i]}_0\}_{i=1}^N$
\end{algorithmic}
\end{algorithm}

Finally, we highlight several additional considerations:
\begin{itemize} 
\item \textbf{Potential Lack of Diversity:} In practice, SMC-generated samples may suffer from sample collapse (i.e., generating the same sample in one batch) for the purpose of reward maximization. This occurs because, as $\alpha \to 0$, the effective sample size approaches $1$. Another challenge with SMC-based methods for reward maximization is that the effective sample size does not necessarily guarantee true diversity among the samples. As a result, even when the effective sample size is controlled, the generated samples may still lack sufficient diversity.
\item \textbf{Eliminating Inferior Samples via Global Interaction:} This algorithm involves interactions between batches, allowing the removal of low-quality particles to concentrate resources on more promising ones. A variant of this strategy has also proven effective in autoregressive language models \citep{zhang2024accelerating}. 
\end{itemize}

\subsection{Value-Based Importance Sampling}\label{sec:SVDD}

Next, we explain a simple value-based importance sampling approach called SVDD, proposed by \citet{li2024derivative}. For this purpose, we first provide an intuitive overview. This method is iterative in nature. Hence, suppose we are at time $t$ and we have $N$ samples, $\{x^{[i]}_{t}\}_{i=1}^N$, with uniform weights: $1/N\sum_{i=1}^N \delta_{x^{[i]}_{t}}. $
Following a proposal distribution $q_{t-1}(\cdot | x^{[i]}_{t})$ (e.g., pre-trained models), we generate $M$ samples $\{ x^{[i,j]}_{t-1}\}_{j=1}^M$ for each $x^{[i]}_{t}$. Ideally, we want to sample from the optimal policy \eqref{eq:soft}. To approximate this, based on importance sampling, we consider the following weighted empirical distribution: 
\begin{align*}
     \frac{1}{N}\sum_{i=1}^N \sum_{j=1}^M \frac{w^{[i,j]}_{t-1}  }{\sum_{j=1}^M w^{[i,j]}_{t-1}}\delta_{ x^{[i,j]}_{t-1}}, \quad w^{[i,j]}_{t-1} = \frac{p^{\star}_{t-1}(x^{[i,j]}_{t-1}|x^{[i]}_{t} ) }{q_{t-1}(x^{[i,j]}_{t-1}|x^{[i]}_{t} )} =  \frac{ p^{\pre}_{t-1}(x^{[i,j]}_{t-1}|x^{[i]}_{t} )\exp(v_{t-1}(x^{[i,j]}_{t-1})/\alpha) }{q_{t-1}(x^{[i,j]}_{t-1}|x^{[i]}_{t} ) \exp(v_{t-1}(x^{[i]}_{t})/\alpha )   }. 
\end{align*}
Since $\exp(v_{t-1}(x^{[i]}_{t})/\alpha )$ in the above remains constant for all $j \in [1,\cdots,M]$, the weight simplifies to
\begin{align}\label{eq:correct_weight}
    w^{[i,j]}_{t-1} = \frac{ p^{\pre}_{t-1}(x^{[i,j]}_{t-1}|x^{[i]}_{t} )\exp(v_{t-1}(x^{[i,j]}_{t-1})/\alpha) }{q_{t-1}(x^{[i,j]}_{t-1}|x^{[i]}_{t} ) }. 
\end{align}
However, repeatedly using this empirical distribution increases the particle size to $O(N^M)$, making it computationally prohibitive. Therefore, we sample to maintain a fixed batch size:
\begin{align*}
    \frac{1}{N} \sum_{i=1}^N \delta_{x^{[i]}_{t-1}}, x^{[i]}_{t-1}:=  x^{[\zeta_i]}_{t-1}, \zeta_i \sim \mathrm{Cat}\left [\left\{   \frac{w^{[i,j]}_{t-1}  }{\sum_{k=1}^M w^{[i,k]}_{t-1}} \right\}_{j=1}^M \right ]. 
\end{align*}
Finally, the complete algorithm is summarized in \pref{alg:decoding2}.

\begin{algorithm}[!th]
\caption{Value-Based Importance Sampling (SVDD)~\citep{li2024derivative}}\label{alg:decoding2}
\begin{algorithmic}[1]
     \STATE {\bf Require}: Estimated (potentially \textbf{non-differentiable}) soft value function $\{\hat v_t\}_{t=T}^0$ (\pref{sec:method_value}), pre-trained diffusion models $\{p^{\pre}_t\}_{t=T}^0$, hyperparameter $\alpha \in \mathbb{R}$, proposal distribution $\{q_t\}_{t=T}^0$, batch size $N$, duplication size $M$
     \FOR{$t \in [T+1,\cdots,1]$}
      \STATE \textbf{Importance sampling step:} For $i\in [1,\cdots,N]$, get $M$ samples from pre-trained polices $\{x^{[i,j]}_{t-1}\}_{j=1}^{M} \sim q_{t-1}(\cdot| x^{[i]}_{t}) $. For each $j\in [1,\cdots,M]$, calculate  $$ w^{[i,j]}_{t-1}:=  \exp(\hat v_{t-1}(x^{[i,j]}_{t-1})/\alpha)\times \frac{p^{\pre}_{t-1}(x^{[i,j]}_{t-1}| x^{[i]}_{t}) }{q_{t-1}(x^{[i,j]}_{t-1}| x^{[i]}_{t})}.$$ \label{line:select3}
        \STATE \textbf{Local resampling step:} $$ \forall i\in [1,\cdots,N];\quad x^{[i]}_{t-1}    \leftarrow  x^{[i, \zeta^{[i]}_{t-1} ]}_{t-1},\quad \zeta^{[i]}_{t-1}  \sim \mathrm{Cat}\left ( \left \{\frac{w^{[i,j]}_{t-1}}{\sum_{k=1}^{M} w^{[i,k]}_{t-1} } \right \}_{j=1}^{M} \right).$$ \label{line:select}
     \ENDFOR
  \STATE {\bf Output}: $\{x^{[i]}_0\}_{i=1}^N$
\end{algorithmic}
\end{algorithm}

\paragraph{Differences Between SMC-Based Methods and Value-Based Importance Sampling (SVDD).}

While SVDD and SMC share similarities, there are two key differences. First, while the sampling in SMC is performed among the whole batch (i.e., global interaction happens), sampling in SVDD is performed for each sample within a batch independently, with no interaction between samples. Second, the weight definition differs: in SMC, the weight includes $\exp(v_t(\cdot))$  in the denominator, while in SVDD, it does not, as normalization occurs independently for each sample, as mentioned in \pref{eq:correct_weight}. 

{ Readers may wonder which approach to use. When the goal is alignment (i.e., reward maximization) in \pref{sec:alignment} (with small $\alpha$), SVDD is more suitable since SMC may suffer from mode collapse. For conditioning tasks in \pref{sec:conditioning} (with moderate $\alpha$), the comparison becomes more nuanced.} As noted in \pref{sec:SMC}, SMC benefits from eliminating inferior samples through global interaction, making it advantageous in certain cases. 

\subsection{Nested-SMC-Based Guidance}

\begin{figure}[!th]
    \centering
    \includegraphics[width=0.8\linewidth]{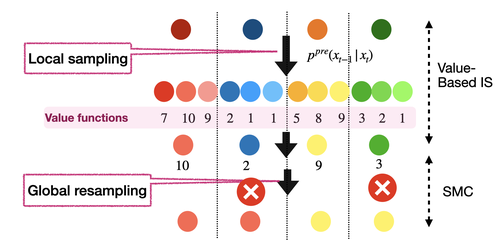}
    \caption{Intuition Behind Nested-SMC-Based Guidance. The algorithm comprises two components: local sampling, which parallels value-based IS sampling, and global resampling, which mirrors SMC-based guidance. }
    \label{fig:nested-SMC}
\end{figure}

As discussed above, the SMC-based approach and the value-based importance sampling approach each have distinct advantages. A hybrid method, referred to as nested-SMC (or nested-IS) in computational statistics \citep[Algorithm 5]{naesseth2019elements}, may combine the strengths of both approaches. This method is outlined in \pref{alg:nested_IS}, and the intuition is outlined in \pref{fig:nested-SMC}. Each step involves two processes: the first is local sampling, resembling value-based importance sampling, and the second is global resampling, characteristic of SMC-based guidance. By incorporating these two elements, nested-IS-based guidance can be more effectively tailored for optimization, allowing the elimination of suboptimal samples through global resampling.

\begin{algorithm}[!th]
\caption{Nested-IS-Based Guidance}\label{alg:nested_IS}
\begin{algorithmic}[1]
     \STATE {\bf Require}: Estimated (potentially \textbf{non-differentiable}) soft value function $\{\hat v_t\}_{t=T}^0$ (\pref{sec:method_value}), pre-trained diffusion models $\{p^{\pre}_t\}_{t=T}^0$, hyperparameter $\alpha \in \mathbb{R}$, proposal distribution $\{q_t\}_{t=T}^0$,  batch size $N$, duplication size $M$
     \FOR{$t \in [T+1,\cdots,1]$}
       \STATE \textbf{Importance sampling step:}  For $i \in [1,\cdots,N]$, get $M$ samples from pre-trained polices $\{x^{[i,j]}_{t-1}\}_{j=1}^{M} \sim q_{t-1}(\cdot| x^{[i]}_{t}) $. And for each $j\in [1,\cdots,M]$, and calculate  $$  w^{[i,j]}_{t-1}:=  \exp(\hat v_{t-1}(x^{[i,j]}_{t-1})/\alpha)\times \frac{p^{\pre}_{t-1}(x^{[i,j]}_{t-1}| x^{[i]}_{t}) }{q_{t-1}(x^{[i,j]}_{t-1}| x^{[i]}_{t})}.$$ \label{line:select3}
        \STATE \textbf{Local resampling step:} $ \{ x^{[i]}_{t-1} \}_{i=1}^N   \leftarrow  x^{[i, \zeta^{[i]}_{t-1} ]}_{t-1} $ where $ \zeta^{[i]}_{t-1}  \sim \mathrm{Cat}\left ( \left \{\frac{w^{[i,j]}_{t-1}}{\sum_{k=1}^{M} w^{[i,k ]}_{t-1} } \right \}_{j=1}^{M} \right),\,$ \label{line:select4}
    \STATE \textbf{Calculate global weight (normalizing constant):}  
$ w^{[i]}_{t-1}=1/M \sum_{j=1}^M  w^{[i,j]}_{t-1}$
    \STATE \textbf{Global resampling step:} Resample by selecting new indices with replacement:  \\ 
       $$\{x^{[i]}_{t-1}\}_{i=1}^N  \leftarrow \{x^{\zeta^{[i]}_{t-1}}_{t-1}\}_{i=1}^N,\quad \{\zeta^{[i]}_{t-1}\}_{i=1}^N \sim \mathrm{Cat}\left ( \left \{\frac{w^{[i]}_{t-1}}{\sum_{k=1}^N w^{[k]}_{t-1} } \right \}_{i=1}^N\right).$$
     \ENDFOR
  \STATE {\bf Output}: $\{ x^{[i]}_0\}_{i=1}^N$
\end{algorithmic}
\end{algorithm}

\subsection{Beam Search for Reward Maximization}\label{sec:beam}

Now, consider a special case of reward maximization where it is natural to set $\alpha = 0$. In this scenario, the SVDD algorithm (\pref{alg:decoding2}) simplifies to \pref{alg:decoding3}, where the selected index corresponds to the one that maximizes the soft value functions.

This algorithm can also be viewed as a beam search guided by soft value functions. Specifically, multiple nodes are expanded according to the proposal distributions, and the best node is selected based on its value function. While this process may seem greedy, it is not, as soft value functions theoretically serve as look-ahead mechanisms, predicting future rewards from intermediate states.

However, the theoretical foundation of this approach relies on the assumption of perfect soft value function access. In practice, approximation errors may arise, and in certain cases, a deeper search might yield additional benefits. We will later explore deeper search algorithms, such as Monte Carlo Tree Search (MCTS), in \pref{sec:MCTS}.

\begin{algorithm}[!th]
\caption{Beam Search with Soft Value Functions \citep{li2024derivative}}\label{alg:decoding3}
\begin{algorithmic}[1]
     \STATE {\bf Require}: Estimated (potentially \textbf{non-differentiable}) soft value function $\{\hat v_t\}_{t=T}^0$ (refer to \pref{sec:method_value}), pre-trained diffusion models $\{p^{\pre}_t\}_{t=T}^0$, proposal distribution $\{q_t\}_{t=T}^0$, batch size $N$, duplication size $M$
     \FOR{$t \in [T+1,\cdots,1]$} 
       \STATE  Get $M$ samples from pre-trained polices $\{x^{[i,j]}_{t-1}\}_{j=1}^{M} \sim p^{\pre}_{t-1}(\cdot| x^{[i]}_{t}) $ 
        \STATE Update $ x^{[i]}_{t-1}   \leftarrow  x^{[i, \zeta^{[i]}_{t-1}] }_{t-1} $ where  $ \textstyle \zeta^{[i]}_{t-1} :=  \argmax_{j \in [1,\cdots,M]} \hat v_{t-1}(x^{[i,j]}_{t-1}).$  \label{line:select}
     \ENDFOR
  \STATE {\bf Output}: $\{ x^{[i]}_0\}$
\end{algorithmic}
\end{algorithm}

\subsection{Selecting Proposal Distributions} \label{sec:choice}

Selecting the appropriate proposal distributions is an important decision for the methods introduced so far from \pref{alg:SMC} to \pref{alg:decoding3}. We outline three fundamental options below.

\paragraph{Pre-Trained Diffusion Polices.} The simplest option is to use the policy from the pre-trained diffusion model.

\paragraph{Derivative-Based Guidance.} Alternatively, derivative-based guidance from \pref{sec:derivative} can be employed, as demonstrated in \citet{wu2024practical,phillips2024particle} with differentiable value function models. Even if the original feedback is non-differentiable and constructing differentiable models is non-trivial, this approach may still outperform pre-trained diffusion models, as the differentiable models can still retain meaningful reward signals.

{\paragraph{Fine-Tuned Policies.}  As we will discuss in \pref{sec:fine_tuning}, when we apply distillation or repeat distillation and inference-time techniques, we can use policies from fine-tuned models as enhanced proposal distributions. }

\section{Derivative-Based Guidance in Continuous Diffusion Models}\label{sec:derivative}

We have introduced the derivative-free inference-time technique. In this section, we focus on classifier guidance \citep{dhariwal2021diffusion,song2021score}, a standard derivative-based method in continuous diffusion models. We first provide the intuition underlying the algorithm's derivation, followed by its formalization within a continuous-time framework. Finally, we propose an algorithm designed for Riemannian diffusion models, which are extensively used in protein srtucture generation.

{\newedit \begin{AIbox}{Takeaway}
Classifier guidance, which adds gradients of value functions at inference time, is derived as an algorithm to approximate the soft optimal policy when discretization errors are negligible. This is formulated within the continuous-time framework using Doob's transform. For Riemannian diffusion models, this algorithm can be seamlessly extended by incorporating Riemannian gradients instead of standard Ecludiain gradients.
\end{AIbox}
} 

\subsection{Intuitive Derivation of Classifier Guidance} \label{sec:intuition}

In this subsection, we derive classifier guidance as a Gaussian policy that approximates the soft-optimal policy. 

First, recalling the form of pre-trained policies in diffusion models over Euclidean space, specifically
\eqref{eq:score_rep} with score parametrization, let us denote the pre-trained model to be $$
p^{\pre}_{t-1}(\cdot \mid x_t) = \Ncal( \rho^{\pre}(x_t,t)  ;\sigma^2_t \rI),\quad \rho^{\pre}(x_t,t) := x_t + (\delta t) \bar g(x_t,t)
$$ 
where
\begin{align*}
  \bar g(x_t,t):=  0.5 x_t  + \widehat{\nabla \log p}(x_t;\theta_{\pre}),\quad \sigma^2_t=(\delta t). 
\end{align*}
Here, $(\delta t)$ is assumed to be small. { Substituting this expression into the optimal policy form in \eqref{eq:optimal_policy} yields } 
\begin{align}\label{eq:original_before}
  \exp\left (\frac{v_{t-1}(x_{t-1})}{\alpha} + \frac{\|x_{t-1}-\rho^{\pre}(x_t,t)\|^2_2}{2\sigma^2_t} \right),      
\end{align}
up to normalizing constant. {However, directly sampling from this policy is challenging; therefore, we consider approximating it.  }

A natural approach is to approximate this policy with a Gaussian distribution. To achieve this, we apply a Taylor expansion:
\begin{align*}
  \exp\left ( \frac{v_t(x_t) + \nabla v_{t}(x_{t})\cdot (x_{t-1}-x_{t}) + O(\|x_{t-1}-x_t\|^2_2)  }{\alpha} \right)  \times \exp\left (\frac{\|x_{t-1}-x_t- (\delta t) \bar g(x_t,t) \|^2_2}{2\sigma^2_t} \right).        
\end{align*}
Since $\sigma^2_t$ is much smaller than $\alpha$ (as $\sigma^2_t$ scales with $(\delta t)$), we can ignore the term $O(\|x_{t-1} - x_t\|^2_2 / \alpha)$. Therefore, the expression simplifies to 
\begin{align*}
   \exp\left (\frac{\|x_{t-1}-x_t - \tilde \rho(x_t,t)\|^2_2}{2\sigma^2_t} \right), \quad  \tilde \rho(x_t,t)=\rho^{\pre}(x_t,t)+\frac{\sigma^2_t \nabla v_{t}(x_t) }{\alpha}. 
\end{align*}
Thus, the original policy in \eqref{eq:original_before} is approximated as a Gaussian distribution.

Based on this observation, the complete algorithm is summarized in \pref{alg:deri}, where the gradient of the value function is incorporated at inference time.

\begin{algorithm}[!th]
\caption{Classifier Guidance in Continuous Diffusion Models \citep{dhariwal2021diffusion,song2021score} }\label{alg:deri}
\begin{algorithmic}[1]
     \STATE {\bf Require}: Estimated (\textbf{differentiable}) soft value functions $\{\hat v_t\}_{t=T}^0$ (refer to \pref{sec:method_value}), pre-trained diffusion models $\{p^{\pre}_t\}_{t=T}^0$, hyperparameter $\alpha \in \mathbb{R}$
     \FOR{$t \in [T+1,\cdots,1]$}
        \STATE $$ x_{t-1} \sim \Ncal\left (\rho^{\pre}(x_t,t)+\frac{\sigma^2_t \nabla \hat v_{t}(x_t) }{\alpha},\sigma^2_t \right) $$
     \ENDFOR
  \STATE {\bf Output}: $x_0$
\end{algorithmic}
\end{algorithm}

\subsection{Derivative-Free Guidance Versus Classifier Guidance}

The critical assumption in classifier guidance is that accurate differentiable value function models can be constructed with respect to the inputs. A straightforward scenario for building such models is an inpainting task, where classifier guidance performs effectively. However, in many molecular design tasks, this assumption may not hold, as discussed in the introduction and \pref{sec:derivative_free}. In such cases, derivative-free guidance becomes advantageous, as these methods do not require differentiable rewards or value function models. 

It is also worthwhile to note these two approaches (derivative-free guidance and classifier-guidance) can still be combined by employing classifier guidance as a proposal distribution in derivative-free methods, potentially with different value functions. Specifically, even if the differentiable models used in classifier guidance are not fully accurate, they can serve as proposal distributions while SMC-based guidance or value-based sampling leverages more precise non-differentiable value function models, as we discuss in \pref{sec:choice}.

\subsection{Continuous-Time Formalization via Doob transform}\label{sec:formalization_conti}

In this section, we formalize the intuitive derivation of classifier guidance in \pref{sec:intuition}. For this purpose, we explain the continuous-time formulation of diffusion models first.

\subsubsection{Preparation}

We begin by outlining the fundamentals of diffusion models within the continuous-time framework. For further details, refer to \citet[Section 1.1.1]{uehara2024understanding} or many other reviews \citep{tang2024score}. The training process can be summarized as follows: (1) defining the forward SDE and (2) learning the time-reversal SDE by estimating the score functions.

\paragraph{Forward and Time-Reversal SDE.}

We first introduce a forward stochastic differential equation (SDE) from $t \in [0,T]$. A widely used example is the variance-preserving (VP) process:
\begin{align}\label{eq:reference}
  t \in [0,T];  dx_t = -0.5 x_t dt +  dw_t, x_0 \sim p^{\pre}(x), 
\end{align} 
where $dw_t$ denotes standard Brownian motion. Two key observations are:
\begin{itemize}
    \item As $T$ approaches $\infty$, the limiting distribution converges to $\Ncal(0,\rI)$. 
    \item The time-reversal SDE \citep{anderson1982reverse}, which preserves the marginal distribution, is given by: 
    \begin{align}\label{eq:time_reversal}
    t \in [0,T];    dz_t = [0.5z_t + \nabla \log q_{T-t}(z_t)]dt + dw_t. 
    \end{align}
    Here, $q_t\in \Delta(\RR^d)$ denotes the marginal distribution at time $t$ induced by the forward SDE \eqref{eq:reference}. Notably, the marginal distribution of $z_{T-t}$ is the same as that of $x_t$ induced by the forward SDE.  Note the notation $t$ is reversed relative to the forward SDE in \eqref{eq:reference}.
    
\end{itemize}

These observations suggest that with sufficiently large $T$, starting from $\Ncal(0,\rI)$ and following the time-reversal SDE \eqref{eq:time_reversal}
, we can sample from the data distribution (i.e., $p^\pre$) at terminal time $T$. A key challenge remains in learning the score function $\nabla \log q_{T-t}(z_t)$. In diffusion models, the primary objective is to estimate this score function. For such training methods, refer to Example~\ref{exa:continuous}. Our work assumes the availability of a pre-trained model $ s(z_{t},T-t; \theta_{\pre})$, that predicts $\nabla \log q_{T-t}(z_t)$, fixed after pre-training. 

\subsubsection{Doob Transform}  

We now proceed to derive classifier guidance more rigorously. Consider a pre-trained model represented by
 \begin{align}\label{eq:pre_trained_SDE}
    t \in [0,T];    dz_t = [0.5z_t + s(z_{t},T-t; \theta_{\pre}) ]dt + dw_t, \quad z_0\sim \delta_{z^\mathrm{ini}_0}.    
\end{align}
We denote the resulting distribution as $p^{\pre}$. The following theorem is instrumental in deriving classifier guidance.

\begin{Thmbox}{}
\begin{theorem}[Doob Transform]\label{thm:doob}
For a value function: $v_t(\cdot) = \log \EE_{\theta^{\pre}}[\exp(r(z_T))|z_t=\cdot]$ where the expectation $\EE_{\theta^{\pre}}[\cdot]$ is  induced by the pre-trained model \eqref{eq:pre_trained_SDE}, the distribution induced by the SDE: 
\begin{align}\label{eq:optimal_SDE}
 t \in [0,T];    dz_t = [0.5z_t + \{ s(z_{t},T-t; \theta_{\pre})+ \nabla \log v_t(z_t)  \}]dt + dw_t
\end{align}
is a target distribution, proportional to $\exp(r(x))p^{\pre}(x)$. 
\end{theorem}
\end{Thmbox}

This theorem implies that, with standard Euler-Maruyama discretization, classifier guidance can be formally derived as in \pref{alg:deri} (when $\alpha=1$) such that we can sample from the target distribution $\exp(r(x))p^{\pre}(x)$. Here, we remark that $v_t(\cdot) = \log \EE_{\theta^{\pre}}[\exp(r(z_T))|z_t=\cdot]$ in the theorem serves as the continuous-time analogue of the value function, which we introduce in \pref{eq:soft}.

The Doob transform is a celebrated result in stochastic processes (e.g., \citet[Chapter 3]{chetrite2015nonequilibrium}). The connection between classifier guidance and the Doob transform has been highlighted in works such as \citet{zhao2024adding,denker2024deft}. Furthermore, \citet{uehara2024finetuning,denker2024deft} demonstrated that, in the context of RL-based fine-tuning, the optimal control that maximizes entropy-regularized rewards coincides with the SDE given in \eqref{eq:optimal_SDE}.

\begin{remark}
When the initial distribution $p^\mathrm{ini}$ from pre-trained models is stochastic, we technically need to change the initial distribution. More specifically, we need to set the initial distribution proportional to  $
    \exp(v_0(x))p^{\mathrm{ini}}(x). $
For details, refer to \citet{uehara2024finetuning}. 
\end{remark}

\subsection{Guidance in Riemannian Diffusion Models}

We now extend the discussion from Euclidean spaces to Riemannian manifolds \citep{de2022riemannian,yim2023se,chen2023riemannian}. To do so, we first provide a concise introduction to Riemannian manifolds, focusing specifically on the special orthogonal group (SO(3)). This is because SE(3) (i.e., $SO(3) \otimes \mathbb{R}^3$) is commonly employed in protein conformation generation to efficiently model the 3D coordinates of the protein backbone \citep{yim2023se,watson2023novo}. Subsequently, we describe the corresponding classifier guidance method. For a more detailed introduction to Riemannian manifolds, we refer readers to \citep{lee2018introduction}. 

\subsubsection{Primer: Riemannian Manifolds}

We denote a $d$-dimensional Riemannian submanifold embedded in $\RR^m$ by $\Mcal$. The manifold is a space that locally resembles Euclidean space, and is formally characterized by a local coordinate chart $\phi: \Mcal \to \RR^d$ and its inverse $\psi$. A key concept in a manifold is the tangent space, which represents the space of possible velocities for a particle moving along the manifold. For each point $x \in \Mcal$, the tangent space $\Tcal_x\Mcal$ is formally defined as the space spanned by the column vectors of the Jacobian $d\psi/d\tilde x |_{\tilde x=\phi(x)}$. A Riemannian manifold is then defined as a manifold equipped with a specific metric $g: \Tcal_x\Mcal \times \Tcal_x\Mcal \to \RR$, often denoted by $\langle\cdot,\cdot\rangle_g$. An important example in protein design is the well-known SO(3) group.

\renewcommand\thmcontinues[1]{Continued}

\begin{Exabox}{}
\begin{example}[SO(3)]
Consider the set of $3 \times 3$ orthogonal matrices, denoted by $SO(3)$. While the intrinsic dimension $d$ is 3, this is embedded in $\RR^9$. The associated tangent space at $x$ (i.e., $\Tcal_{x}(SO(3))$) consists of skew-symmetric matrices: $\{xA:  A^{\top} = -A\}$ where the Riemannian metric is defined using the Frobenius inner product:
$\langle A_1,A_2\rangle_{g}= \mathrm{Tr}(A^{\top}_1 A_2).$ Specifically, the tangent space corresponds to 
\begin{align*}
    \left \{x[v_1, v_2, v_3 ]_{\times}: v_1 \in \RR, v_2 \in \RR, v_3 \in \RR \right\}, \quad [v]_{\times}\textstyle :=
    {\tiny
    \begin{pmatrix}
        0 & -v_3 & v_2 \\
        v_3 & 0 & -v_1 \\
        -v_2 & v_1 & 0. 
    \end{pmatrix}.  } 
\end{align*} 
This can be derived by introducing a curve $x(t) \in SO(3)$ such that $x(0)=x$, and computing the gradient of $x^{\top}(t)x(t)=I$. The gradient $\nabla x(0)$ satisfies 
\begin{align*}
    \nabla x(0)^{\top} \cdot x  +  x^{\top} \cdot \nabla x(0)=0. 
\end{align*}
Then, it is clear $\nabla x(0)$ belongs to $\{xA:  A^{\top} = -A\}$. 
\end{example}
\end{Exabox}

We now summarize additional key concepts:
\begin{itemize}
    \item \textbf{Geodesic}: Given a point $x \in \Mcal$ and a velocity $v \in \Tcal_{x} \Mcal$, the geodesic is defined as a trajectory $\gamma(t):\Rcal \to \Mcal$, determined by the initial point $x$ and velocity $v$: 
    \begin{align}\label{eq:curve}
        \gamma(0)=x, \frac{d\gamma(t) }{dt}|_{t=0}=v, 
    \end{align}
    along with the geodesic equation. This trajectory is locally uniquely defined. For example, in Euclidean spaces, it simplifies to $x + tv$. Hence, intuitively, geodesic is the locally shortest path on a Riemmanin manifold. 
    \item \textbf{Exponential map:} Given $x \in \Mcal$, the exponential map is defined as $\Tcal_{x}\Mcal: v \to \gamma(1) \in \Mcal$, where $\gamma(t)$ is the geodesic described above.
    \item \textbf{Logarithmic Map:} The logarithmic map is the inverse of the exponential map.
    \item \textbf{Riemannian Gradient}: The Riemannian gradient of a function $f:\Mcal \to \RR$ at $x$ denoted by $\nabla^{g} f(\cdot)\in \Tcal_x(\Mcal)$ is defined as an element satisfying  
    \begin{align*} 
    \forall v\in \Tcal_{x}\Mcal;\quad \frac{df(\gamma(t))}{dt}|_{{t=0}} = \langle \nabla^{g} f(x),v\rangle_{g}, 
    \end{align*} 
    where $\gamma(t)$ is a curve satisfying \eqref{eq:curve}. This reduces to a standard gradient in a Ecludiain space. 
\end{itemize}

Now, let’s revisit the example of SO(3) to see how these concepts are applied.

\begin{Exabox}{}
\addtocounter{example}{-1}
\begin{example}[continued]
Given a point $x \in SO(3)$ and velocity $v \in \Tcal_{x}(SO(3))$, the exponential map is defined as $x\exp(v)$. Using Rodrigues' rotation formula \citep{lee2018introduction}, it can be simplified to 
{ \begin{align*}
      x\{I + \sin(\alpha)[v/\|v\|_2 ]_{\times} +(1-\cos(\alpha))\{[v/\|v\|_2]_{\times}\}^2  \}, 
\end{align*}
} 
where $\alpha = \|v\|_2$. Then, the Riemannian gradient in $\Tcal_{x}(SO(3))$ is given by :
\begin{align}\label{eq:gradient}
    0.5(\nabla^{\mathrm{Ecu}} f(x) - \{\nabla^{\mathrm{Ecu}} f(x)\}^{\top})  
\end{align}
where $\nabla^{\mathrm{Ecu}}$  is the Euclidean gradient. 
\end{example}
\end{Exabox}

\subsubsection{Classifier Guidance in Riemannian Diffusion Models}

\begin{algorithm}[!th]
\caption{Classifier Guidance in Riemannian Diffusion Models  }\label{alg:deri_rie}
\begin{algorithmic}[1]
     \STATE {\bf Require}: Estimated (\textbf{differentiable}) soft value function $\{\hat v_t\}_{t=T}^0$ (refer to \pref{sec:method_value}), pre-trained diffusion models $\{p^{\pre}_t\}_{t=T}^0$, hyperparameter $\alpha \in \mathbb{R}$ 
     \FOR{$t \in [T+1,\cdots,1]$}
     \STATE Calculate the velocity to proceed:
     \begin{align*}
 \textstyle  \mathrm{vel}_t=(\delta t)\{ s(x_t, T-t; \theta_{\pre}) + \nabla^{g} \hat v_t(x_t)/\alpha\}  + \sqrt{(\delta t)}\epsilon_t, \quad \epsilon_t\sim {\Ncal(0,I_{d})},         
     \end{align*}
     where $\Ncal(0,I_d)$ is Gaussian distribution on a Riemmaninan manifold. 
        \STATE Move from $x_t$ to $x_{t-1}$ along geodesics: $$  \textstyle    x_{t-1}= \exp_{x_t}[\mathrm{vel}_t] $$
     \ENDFOR
  \STATE {\bf Output}: $x_0$
\end{algorithmic}
\end{algorithm}

Now, with the above preparation, the pre-trained model is defined as an SDE on a manifold $\Mcal$:
\begin{align*}
    t \in [0;T];\quad d x_t = s(x_t, T-t; \theta_{\pre})  dt + dw^{\Mcal}_t, 
\end{align*}
where $dw^{\Mcal}_t$ denotes Brownian motion on a Riemannian manifold $\Mcal$. The discretization is given by:
\begin{align*}
    x_{t-1}= \exp_{x_t}[\mathrm{vel}_t], \mathrm{vel}_t=(\delta t)s(x_t, T-t; \theta_{\pre}) + \sqrt{(\delta t)}\epsilon_t, \quad \epsilon_t\sim \Ncal(0,I_{d}),
\end{align*}
where $\epsilon_t$ is a normal distribution on the manifold $\Mcal$. Each step thus consists of two operations: (1) calculating the tangent (velocity) $\mathrm{vel}_t$ in the tangent space at $x_t$ and (2) moving along the geodesic induced by the velocity $\mathrm{vel}_t$, starting at $x_t$. In the Euclidean case, the second step reduces to $x_t +\mathrm{vel}_t$. 

\paragraph{Classifier Guidance.} Now, we extend classifier guidance to Riemannian manifolds. Similar to the Euclidean case, we calculate a Riemannian gradient at each time step during inference and incorporate it into the velocity. The updated velocity becomes:
\begin{align*}
    (\delta t) \{ s(x_t, T-t; \theta_{\pre}) + \underbrace{\nabla^g  v(x_t)/\alpha}_{\textit{Additional\,Riemannian gradient}} \}  + \sqrt{(\delta t)}\epsilon_t. 
\end{align*}
For example, in the case of $SO(3)$, the Riemannian gradient is computed using \eqref{eq:gradient}. 

The complete algorithm is summarized in \pref{alg:deri_rie}. Note this approach can be formalized using Doob's theorem for Riemannian manifolds, as discussed in \pref{sec:formalization_conti}.

\section{Derivative-Based Guidance in Discrete Diffusion Models} \label{sec:derivative_discrete}

We now focus on inference-time techniques specifically designed for discrete diffusion models. In \pref{sec:soft}, we highlighted that exact sampling from the optimal policy is feasible within discrete diffusion models under certain limited scenarios. Here, 
we revisit this point by demonstrating that exact sampling from the optimal policy can be achieved through polynomial-time computation of value functions. Building on this point, we explain derivative-based guidance following the approach of \citet{nisonoff2024unlocking}. Finally, we formalize the discussion within a continuous-time framework \citep{wang2024finetuning}.

\begin{AIbox}{Takeaways}
 In discrete diffusion, while the original action space grows exponentially with respect to token length, the nominal action space is polynomial, allowing optimal policies to be sampled with polynomial computation of value functions. This argument has been formalized using Doob's transform in the continuous-time formulation.

Despite this, polynomial computation remains computationally expensive in many practical scenarios. One approach to mitigate this issue is to employ derivative-based guidance.
\end{AIbox}

\subsection{Exact Sampling in Discrete Diffusion Models}\label{sec:discrete_first}

Our objective in this subsection is to show that sampling from the optimal policy can be achieved with polynomial-time computation of value functions in discrete diffusion models. 
As a preliminary step, we demonstrate that the effective action space in the pre-trained policy scales \emph{polynomially} with respect to token length, rather than \emph{exponentially}. We then extend this argument to show that the effective action space in the optimal policy also scales polynomially. This leads to the conclusion that sampling from the optimal policy is computationally feasible using polynomial computation, avoiding the need for exponential computation of value functions.

\paragraph{Closer look at policies from pre-trained models.} Our first objective here is to demonstrate that the effective action space in discrete diffusion models is $LK$ rather than $K^L$, where $L$ is the token length and $K$ is the vocabulary size.

To begin, recall that the pre-trained policy in diffusion models is expressed as:
\begin{align}\label{eq:original}
    \prod_{l=1}^L p^{\pre}_{t-1}(x^{l}_{t-1} | x^{1:L}_{t}),\,\, \textit{where}\,\,  p^{\pre}_{t-1}(x^{l}_{t-1} | x^{1:L}_{t}):=\mathrm{I}(x^l_t=x^l_{t-1}) +     Q^{\pre}_{x^l_{t-1}, x^{1:L}_{t} }(t)   (\delta t),
\end{align}
where $(\delta t)$ is a discretization step. For example, in maskd diffusion models mentioned (Example~\ref{exa:discrete}), 
\begin{align*}
     Q^{\pre}_{x^l_{t-1}, x^{1:L}_{t} }(t):=  
     \begin{cases}
            \frac{\alpha_{t-1} - \alpha_t  }{1-\bar \alpha_t}\hat x_0(x_t;\theta^{\pre})\quad \textit{when}\,\,x^l_{t-1}\neq x^l_t, \, x^l_t=\textbf{Mask} \\
            -\sum_{z \neq x^{1:L}_{t} }Q^{\pre}_{z, x^{1:L}_{t} }(t) \,\,\textit{when}\,x^l_{t-1}= x^l_t,\,x^l_t=\textbf{Mask}, \\
            0\quad \textit{when}\,\,x^l_t\neq \textbf{Mask}. 
     \end{cases}
\end{align*}
{ At first glance, evaluating the optimal policy based on the pre-trained policy \eqref{eq:original}, i.e., 
\begin{align*}
    \exp(v(x^{1:L}_{t-1}))\prod_{l=1}^L p^{\pre}_{t-1}(x^{l}_{t-1} | x^{1:L}_{t})
\end{align*}
might seem computationally prohibitive}, as it appears to require $O(L^K)$ evaluations of the value functions. However, it can actually be computed using only $O(LK)$ operations. The key idea lies in identifying an asymptotically equivalent policy, as described below.
\begin{itemize}
    \item \textbf{Case 1: A single Token Change.} Consider the scenario where only a single token (position $c \in {1, \dots, L}$) changes in $x^l_{t-1}$. The transition probability is: 
\begin{align*}
     p^{\pre}_{t-1}(x^{1:L}_{t-1} \mid x^{1:L}_t)= Q^{\pre}_{x^c_{t-1}, x^{1:L}_{t} }(t) (\delta t)  \times \prod_{l \neq c} \left\{ 1 +Q^{\pre}_{x^l_{t-1}, x^{1:L}_{t} }(t) (\delta t)  \right \}. 
\end{align*}
By some algebra, we can approximate this as:
\begin{align*}
         p^{\pre}_{t-1}(x^{1:L}_{t-1} \mid x^{1:L}_t)=\underbrace{ Q^{\pre}_{x^c_{t-1}, x^{1:L}_{t} }(t)(\delta t)}_{O(\delta t)\mathrm{term\,(first\,order)} } + O((\delta t)^2). 
\end{align*}
\item \textbf{Case 2: Multiple Token Changes.}  When multiple tokens change in $x^l_{t-1}$, similar calculations yield $p^{\pre}_{t-1}(x^{1:L}_{t-1} \mid x^{1:L}_t)=O((\delta t)^{\mathrm{Ch}(x^{1:L}_{t-1},x^{1:L}_t)})$  where $\mathrm{Ch}$ denotes the number of token changes between  $x^{1:L}_{t-1}$ and $x^{1:L}_t$, defined as $\mathrm{Ch}(x_{t-1},x_t) := \|x_{t-1} - x_t\|_1 / 2$.
\end{itemize}

Summarizing the above, ignoring second-order terms, the resulting asymptotically equivalent policy becomes:
\begin{align}\label{eq:pre_trained_discrete}
    p^{\pre}_{t-1}(x^{1:L}_{t-1} \mid x^{1:L}_t)=\begin{cases}
       0, \quad  \mathrm{Ch}(x^{1:L}_{t-1},x^{1:L}_t) \geq   2,  \\
      Q^{\pre}_{x^{1:L}_{t-1},x^{1:L}_t}(t)  (\delta t),\quad  \mathrm{Ch}(x^{1:L}_{t-1},x^{1:L}_t) = 1,  \\ 
     1 +  Q^{\pre}_{x^{1:L}_{t-1},x^{1:L}_t}(t)(\delta t), \quad  \mathrm{Ch}(x^{1:L}_{t-1},x^{1:L}_t) = 0, 
    \end{cases}
\end{align}
where $ Q^{\pre}_{x^{1:L}_{t-1},x^{1:L}_t}(t)$ in the second line implicitly means $Q^{\pre}_{x^{c}_{t-1},x^{1:L}_t}(t)$ where $c$ is the changed token place, and the generator in the third line is 
$$ 
Q^{\pre}_{x^{1:L}_{t},x^{1:L}_t}(t) = -\sum_{\{ x'_{t-1}: x'_{t-1}\neq x^{1:L}_t, \mathrm{Ch}(x'_{t-1},x^{1:L}_t)=1 \} }  Q^{\pre}_{x'_{t-1},x^{1:L}_t}(t). 
$$
Crucially, since the effective action space in $x_{t-1}$ is $LK$ (instead of exponential in $L$), the computational cost of the above summation only requires $LK$ operations, ensuring polynomial complexity.

\paragraph{Sampling from the Optimal Policy.} Now, let us revisit the formulation of the optimal policy. For simplicity, we denote  $x^{1:L}_{t-1}$ as $x_{t-1}$. Substituting \eqref{eq:pre_trained_discrete} into the form of the optimal policy \eqref{eq:optimal_policy}, we obtain:
\begin{align*}
    p^{\star}_{t-1}(x_{t-1} \mid x_t)=\begin{cases}
       0, \quad \mathrm{Ch}(x_{t-1}, x_{t}) \geq  2, \\
      \frac{\exp(v_{t-1}(x_{t-1})) Q^{\pre}_{x_{t-1},x_t}(t)  (\delta t)}{\exp(v_{t-1}(x_{t}))\{1 +  Q^{\pre}_{x_{t},x_t}(t)(\delta t)\}  + \sum_{x'_{t-1}} \exp(v_{t-1}(x'_{t-1})) Q^{\pre}_{x'_{t-1},x_t}(t)  (\delta t)   }, 
      \quad  \mathrm{Ch}(x_{t-1}, x_{t}) = 1, \\ 
    \frac{ \exp(v_{t-1}(x_{t}))\{1 +  Q^{\pre}_{x_{t-1},x_t}(t)(\delta t)  \} }{\exp(v_{t-1}(x_{t}))\{1 +  Q^{\pre}_{x_{t},x_t}(t)(\delta t)\}   + \sum_{x'_{t-1}} \exp(v_{t-1}(x'_{t-1})) Q^{\pre}_{x'_{t-1},x_t}(t)  (\delta t)   }, \quad  \mathrm{Ch}(x_{t-1}, x_{t}) = 0.
    \end{cases}
\end{align*}
This formulation may seem complex, so let’s simplify it.

For the case when $\mathrm{Ch}(x_{t-1}, x_{t}) = 1$, the expression simplifies to:
\begin{align*}
       \frac{\exp(v_{t-1}(x_{t-1})) Q^{\pre}_{x_{t-1},x_t}(t)  (\delta t)}{\exp(v_{t-1}(x_{t})) } + O((\delta t)^2). 
\end{align*}

Ignoring higher-order terms $O((\delta t)^2)$, we can further streamline the formulation. The complete algorithm is summarized in  \pref{alg:claasify_dscrete} \citep{nisonoff2024unlocking}.

The key takeaway is that we can again avoid the curse of token length, as the computational cost of evaluating the value function remains $LK$. However, in practical applications, even this computational cost may be prohibitive. To address this, two approximation strategies can be employed:
\begin{itemize}
    \item \textbf{SMC-Based Guidance (\pref{sec:SMC}) or Value-Based Sampling (\pref{sec:SVDD}) }: These methods are more feasible in practice. Notably, even the value-based sampling described in \pref{sec:derivative_free} requires only $M$ evaluations of the value function, a significant reduction compared to $LK$.
\item \textbf{Derivative-Based Approximation \citep{nisonoff2024unlocking,vignac2023digress}}: In the next section, we introduce a further reduction technique by computing derivatives of the value functions, effectively minimizing the $LK$ computation overhead.
\end{itemize}

\begin{remark}[Combination with More Advanced Discretization Methods]
We have employed the most basic discretization method to define policies. However, many state-of-the-art approaches could potentially be applied for policy distillation \citep{campbell2022continuous,ren2024discrete,zhao2024informed}.
\end{remark}

\begin{algorithm}[!th]
\caption{Classifier Guidance in Discrete Diffusion Models \citep{nisonoff2024unlocking} }\label{alg:claasify_dscrete}
\begin{algorithmic}[1]
     \STATE {\bf Require}: Estimated (\textbf{potentially nondifferentiable}) soft value function $\{\hat v_t\}_{t=T}^0$ (refer to \pref{sec:method_value}), pre-trained diffusion models 
     \FOR{$t \in [T+1,\cdots,1]$}
     
        \STATE Sample from the following: 
        \begin{align*}
      p^{\star}_{t-1}(x_{t-1} \mid x_t)=\begin{cases}
          0, \quad  (\mathrm{Ch}(x_{t-1}, x_{t}) \geq  2), \\
       \frac{\exp(\hat v_{t-1}(x_{t-1})) Q^{\pre}_{x_{t-1},x_t}(t)  (\delta t)}{\exp(\hat v_{t-1}(x_{t})) }, \quad    (\mathrm{Ch}(x_{t-1}, x_{t}) =1),  \\
            1-  \sum_{\{ x'_{t-1}: \mathrm{Ch}(x'_{t-1}, x_{t})=1\}  }\frac{\exp(\hat v_{t-1}(x'_{t-1})) Q^{\pre}_{x'_{t-1},x_t}(t)  (\delta t)}{\exp(\hat v_{t-1}(x_{t})) }
            \quad    (\mathrm{Ch}(x_{t-1}, x_{t}) =0).  
      \end{cases}
\end{align*}
     \ENDFOR
  \STATE {\bf Output}: $x_0$
\end{algorithmic}
\end{algorithm}

\subsection{Derivative-Based Guidance in Discrete Diffusion Models}\label{sec:derivative_discrete2}

As we did in \pref{sec:intuition}, we perform a Taylor expansion using a one-hot representation. For the case where $\mathrm{Ch}(x_{t-1}, x_{t})=1$, the expression becomes:
\begin{align*}
   \frac{ \{\exp(v_{t}(x_{t})) +  \nabla \exp(v_{t}(x_{t}))\cdot (x_{t-1}-x_t) + O( \|x_{t-1}-x_t\|^2_2 ) \} }{ \exp(v_{t}(x_{t})) }\times Q^{\pre}_{x_{t-1},x_t}(t)  (\delta t), 
\end{align*}
where $x_t$ is embedded into a \emph{$L\times K$ dimensional one-hot representation}. 
By ignoring the higher-order terms $O( \|x_{t-1}-x_t\|^2_2 )$, this simplifies to:
\begin{align}\label{eq:before_approximation}
   Q^{\pre}_{x_{t-1},x_t}(t)\{ 1  +   \nabla v_{t}(x_{t}) \cdot (x_{t-1}-x_t) \}   (\delta t). 
\end{align}
\citet{nisonoff2024unlocking} proposed this approximation as a practical implementation of classifier guidance in discrete diffusion models. This procedure is summarized in \pref{alg:claasify_dscrete2}. 

\begin{remark} Another asymptotically equivalent formulation of \eqref{eq:before_approximation}  is  $ Q^{\pre}_{x_{t-1},x_t}(t)\exp(\nabla v_{t}(x_{t}) \cdot (x_{t-1}-x_t))  (\delta t)$ noting $\exp(x)\approx 1+x$ when $x$ is small. 
\end{remark}

\begin{algorithm}[!th]
\caption{Classifier Guidance with Taylor Approximation in Discrete Diffusion Models \citep{nisonoff2024unlocking}  }\label{alg:claasify_dscrete2}
\begin{algorithmic}[1]
     \STATE {\bf Require}: Estimated (\textbf{differentiable}) soft value function $\{\hat v_t\}_{t=T}^0$ (refer to \pref{sec:method_value}), pre-trained diffusion models 
     \FOR{$t \in [T+1,\cdots,1]$}
      \STATE { Define the following generator: 
      \begin{align*}
       Q^{\mathrm{new}}_{x^l_{t-1}, x^{1:L}_{t} }(t)&:=   Q^{\pre}_{x_{t-1},x_t}(t) \{1   +   [ \nabla  \hat v_{t}( x^{1:L}_{t})]_{l} \cdot ( x^l_{t-1} -x^l_{t}) \} \,(\textit{if}\, x^l_{t-1}\neq  x^l_{t} ) \\ 
       Q^{\mathrm{new}}_{x^l_{t-1}, x^{1:L}_{t} }(t)   &:= -\sum_{x^l_{t-1} \neq x^l_{t}}  Q^{\mathrm{new}}_{x^l_{t-1}, x^{1:L}_{t} }(t)   \,(\textit{if}\, x^l_{t-1}=  x^l_{t} )
      \end{align*} } 

        \STATE  Sample from 
        \begin{align*}
    \prod_{l=1}^L p^{\mathrm{gui}}_{t-1}(x^{l}_{t-1} | x^{1:L}_{t}),\,\, \textit{where}\,\,\,   p^{\mathrm{gui}}_{t-1}(x^{l}_{t-1} | x^{1:L}_{t}):=\mathrm{I}(x^l_t=x^l_{t-1}) + Q^{\mathrm{new}}_{x^l_{t-1}, x^{1:L}_{t} }(t)   (\delta t). 
\end{align*}
     \ENDFOR
  \STATE {\bf Output}: $x^{1:L}_0$
\end{algorithmic}
\end{algorithm}

However, it is important to note that this practical version could have a potential issue. Most notably, in discrete spaces, formal derivatives cannot be done, rendering the Taylor expansion technically invalid. Consequently, unlike in continuous domains, the derivative-based approach lacks formal guarantees in discrete spaces.

{ \subsection{Continuous-Time Formalization via Doob Transform}}

We now formalize the classifier guidance method for discrete diffusion models in continuous time formulation, as outlined in \pref{sec:discrete_first}. To do so, we first need to understand how discrete diffusion models are framed within the continuous-time formulation. Finally, we formalize the classifier guidance method for diffusion models using the Doob transform in continuous-time Markov chains (CTMC).

\subsubsection{Preparation}

The training of discrete diffusion models follows the same principles as Euclidean diffusion models. We define the forward noising process and aim to learn the denoising process. However, in discrete diffusion models, we must account for the transition from Brownian motion to continuous-time Markov chains (CTMC). Due to this difference, the ``score'' is replaced with the ``ratio''. This substitution is expected, as ratio matching is a well-established method for estimating unnormalized models with discrete data \citep{hyvarinen2007some}, similar to how score matching is widely used for continuous data \citep{hyvarinen2005estimation}.

\paragraph{Forward and Time-Reversal SDE.}

We consider the following family of distributions $q_t \in \RR^{K}$ (a vector summing to 1),  which evolves from $t=0$ to $t=T$ according to a CTMC:
\begin{align*} 
\frac{dq_t}{dt} = Q(t) q_t,\quad p_0\sim p_{\data}, 
\end{align*} 
where $Q(t) \in \RR^{K \times K}$ is the generator. Typically, $q_t$ is designed to converge toward a simple limiting distribution as $t \to T$.  A common strategy is to introduce a \textit{mask} into $\Xcal$ and gradually mask a sequence so that the limiting distribution becomes fully masked \citep{shi2024simplified,sahoo2024simple}. Specifically, these works define the forward masking process so that $q_t \sim \mathrm{Cat}(\bar \alpha_t x_t + (1 - \bar \alpha_t) \mathbf{m})$ as we saw in Example~\ref{exa:discrete}.

Now, we consider the time-reversal CTMC \citep{sun2022score}, which preserves the marginal distribution. This can be expressed as follows: 
\begin{align*}
    \frac{dq_{T-t}}{dt} = \bar Q(T-t)q_{T-t},\quad \bar Q_{y,x}(t) = \begin{cases}
     \textstyle    \frac{q_t(y)}{q_t(x)} Q_{x,y}(t)\,(y \neq x), \\ 
    \textstyle     -\sum_{z\neq x} \bar Q_{z,x}(t)\,(y=x), 
    \end{cases} 
\end{align*}
where $Q_{y,x}(t)$ is a $(y,x)$-entry of a generator $Q(t)$. This formulation implies that if we can learn the marginal density ratio $q_t(y)/q_t(x)$, we can sample from the data distribution at $t=T$ by following the above CTMC governed by $\bar Q(T-t)$. For details on training this ratio, see \citet{lou2023discrete}. In the subsequent discussion, we assume that the pre-training phase is complete and a pre-trained discrete diffusion model is available.

\subsubsection{Doob Transform in CTMC}

We are now prepared to formally derive classifier guidance in discrete diffusion models. Assume we have a pre-trained model characterized by the generator $Q^{\pre}(t)$: 
\begin{align*} \frac{dz_t}{dt} = Q^{\pre}(t)z_t, \quad z_0 \sim \delta(z^\mathrm{ini}_0). \end{align*}
The distribution generated by the above CTMC corresponds to $p^{\pre}$, which characterizes the natural-like design space. Note that time $0$ represents the noise and time $T$ represents the terminal here. 

With the above preparation, we present the following key theorem: Doob transform in CTMC. 

\begin{Thmbox}{}
\begin{theorem}[Doob transform (From Theorem 1, 2 in \citet{wang2024finetuning}) ]\label{thm:doob2}
Define the soft value functions as $v_t(\cdot):=\log \EE_{Q^{\pre}}[\exp(r(z_T))|z_t=\cdot] $ where the expectation is taken with respect to the distribution induced by the pre-trained model. Then, the distribution induced by the CTMC: 
\begin{align}\label{eq:doob_CTMC}
    \frac{dz_t}{dt} = Q^{\star}(t)z_t,\quad  Q^{\star}_{x,y}(t) = Q^{\pre}_{x,y}(t)\exp(\{v_t(y)-v_t(x)\}), 
\end{align}
is the target distribution, proportional to the target distribution $\exp(r(x))p^{\pre}(x)$. 
\end{theorem}
\end{Thmbox}

This theorem indicates that, using standard Euler-Maruyama discretization, we can derive the classifier guidance algorithm introduced in \pref{alg:claasify_dscrete} such that we can sample from the target distribution. Here, we remark that $v_t(\cdot):=\log \EE_{Q^{\pre}}[\exp(r(z_T))|z_t=\cdot] $ is seen as the continuous-time analog of the value function in \pref{eq:soft}. 

The Doob transform for CTMCs is well-established in the literature on the stochastic process (e.g., \citet[Chapter 17]{levin2017markov}; \citet[Chapter 3]{chetrite2015nonequilibrium}). In the context of guidance for discrete diffusion models, \citet{nisonoff2024unlocking} and \citet{wang2024finetuning} introduce this form. Notably, \citet{wang2024finetuning} also proves that, in the context of RL-based fine-tuning, the above CTMC in \eqref{eq:doob_CTMC} maximizes the entropy-regularized reward objective.

\section{Tree Search Algorithms for Alignment}\label{sec:MCTS}

A promising approach for more accurate inference alignment involves leveraging search-based algorithms. As outlined in \pref{sec:beam}, SVDD in \citet{li2024derivative} performs beam search using a value function. More sophisticated algorithms, such as Monte Carlo Tree Search (MCTS) \citep{kocsis2006bandit,silver2016mastering,hubert2021learning,xiao2019maximum}, can also be applied to diffusion models by appropriately defining the search tree. Key aspects to address include determining the depth and width of the search tree, followed by the evaluation of leaf nodes.

Before delving into these aspects in detail, we observe that, unlike in language models \citep{feng2023alphazero,hao2023reasoning}, relatively few studies (e.g., \citet{li2024derivative}) have explored this direction within the context of diffusion models. Given the success of MCTS in molecular generation \citep{yang2017chemts,kajita2020autonomous,yang2020practical,swanson2024generative} in general, we believe that this approach offers considerable potential for molecular design even when using diffusion models. We encourage further research in this area.

\begin{AIbox}{Key Message in Section~\ref{sec:MCTS}} 
Pre-trained diffusion models inherently induce a tree structure that characterizes natural-like designs. Search algorithms, such as Monte Carlo Tree Search (MCTS), can be applied to maximize rewards while effectively utilizing value functions.
\end{AIbox}

\subsection{Defining the Search Tree}

We now discuss how to properly define the search tree.

\paragraph{Original Tree Depth/Width.}
In diffusion models, a natural approach is to set the original tree depth according to the discretization level, typically ranging from $50$ to $1000$. However, the original tree width in diffusion models is considerably larger. For example, in continuous diffusion, the action space lies in a high-dimensional Euclidean space, making exhaustive search computationally prohibitive. Even in masked discrete diffusion models, the tree width is $LK$ (as we mention in \pref{sec:discrete_first}), where $L$ denotes the token length and $K$ the vocabulary size.

\paragraph{Search Tree Depth/Width.} The primary challenge lies in managing the large tree width and depth discussed above. To address this, we must adopt search algorithms capable of strategically limiting node expansion and constraining both the width and depth of the tree. Specifically, starting from the root node (the current state during inference), we limit the tree width and depth by sampling nodes from pre-trained models during expansion, preventing further growth beyond a specified limit. This subsampling strategy, which leverages pre-trained models, has been effectively used in RL like AlphaZero and its variants \citep{hubert2021learning,grill2020monte}, as well as in language models \citep{feng2023alphazero,hao2023reasoning}.

\begin{wrapfigure}{r}{0.5\textwidth}
    \centering
    \vspace{-0.5cm}
\includegraphics[width=0.5\textwidth]{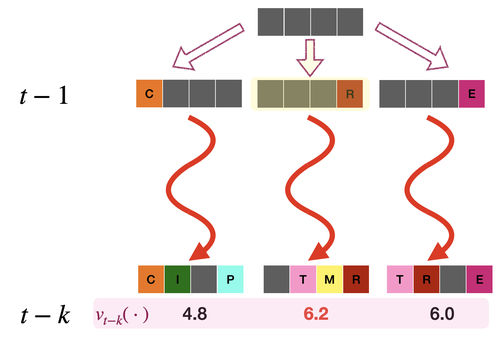}
    \caption{Evaluation of leaf nodes with estimated value functions at $t-k$ rather than $t-1$ by further rolling out pre-trained policies. }
    \label{fig:MCTS}
    \vspace{-0.5cm}
\end{wrapfigure}

\subsection{How to Run Simulations from Leaf Nodes}

While any off-the-shelf search algorithm can be applied to the tree described above, a key decision is how to evaluate the leaf node in the search tree (i.e., the simulation phase in MCTS). To explain it, in the following, let us assume the leaf node is $x_{t-1}$.

The simplest way to evaluate the leaf node is by using a value function approximation, such as the posterior mean approximation in \pref{sec:method_value}, and setting it as $r(\hat{x}_0(x_{t-1}))$. However, as $t$ increases, this approximation generally becomes less accurate. A potentially more effective strategy is to evaluate $r(\hat{x}_0(x_{t-k}))$ after running the pre-trained model for an additional $k$ steps. This approach can yield a more precise estimate, as $t - k$ is closer to 0. The algorithm, combined with beam search, is illustrated in \pref{fig:MCTS}.

However, it is important to note that this $k$-step look-ahead incurs additional computational costs. In the extreme case, running the model for $t$ steps would provide the most accurate evaluation of $r(x_0)$ at time 0 but would be computationally prohibitive. We leave the exploration of this trade-off between accuracy and computational efficiency for future work.

 \section{Editing and Refinement with Diffusion Models}\label{sec:editing}

Thus far, we have focused on inference-time techniques that generate samples from scratch, progressing from a noise state (at $t=T$) to the final state (at $t=0$). In this section, we aim to address two practical scenarios:
\begin{itemize}
\item \emph{Editing:} Particularly in biological sequence design, our task is often to \emph{edit} an existing sequence while preserving its original properties and enhancing its target properties. For instance, in antibody design, the typical number of mutations is limited to $5-10$ \citep{bachas2022antibody}. In such cases, the inference-time techniques discussed so far cannot be directly applied.
\item \emph{Refinement:} Another closely related motivation is the refinement of generated designs. Even when designs are generated from complete noise using inference-time techniques, we may want to \emph{refine} these designs further. Note that many decoding algorithms for BERT-style models have been developed with this objective in mind.
such as Gibbs sampling \citep{wang2019bert,miao2019cgmh}.
\end{itemize}
 In the following sections, we describe how inference-time techniques introduced so far can be adapted for tasks involving editing or refinement.

\begin{AIbox}{Key Message in \pref{sec:editing}}
Inference-time techniques can be applied to the editing or refinement of designs by incorporating mutation steps within an evolutionary algorithm.
\end{AIbox}

\subsection{Iterative Refinement in Diffusion Models} 

\begin{algorithm}[!th]
\caption{Iterative Refinement for Reward Optimization  in Diffusion Models}\label{alg:decoding4}
\begin{algorithmic}[1]
     \STATE {\bf Require}: Estimated soft value function $\{\hat v_t\}_{t=T}^0$ (refer to \pref{sec:method_value}), pre-trained diffusion models $\{p^{\pre}_t\}_{t=T}^0$ and an initial sequence $x^{\langle 0 \rangle }_0$ (the index $\langle \cdot \rangle $ means the number of iteration steps). 
     \FOR{$s \in [0,\cdots,S-1]$} 
       \STATE  Sample $x^{\langle s+1 \rangle }_k$ from  $q_k(\cdot \mid x^{\langle s \rangle }_0)$ where $q_k$ is a pre-defined noising policy from $x_0$ to $x_k$. 
        \STATE Use inference-time techniques introduced so far (methods in \pref{sec:derivative_free}, \pref{sec:derivative}), which combine value function and pre-trained models to obtain $x^{\langle s+1 \rangle }_0$, starting from $x^{\langle s+1 \rangle }_k$
    \STATE (Optional): Filter designs that meet the specified constraints.
     \ENDFOR
  \STATE {\bf Output}: $\{ x^{\langle S \rangle}_0\}$
\end{algorithmic}
\end{algorithm}

Here, we present an editing-type algorithm designed for reward optimization with diffusion models, summarized in \pref{alg:decoding4}.

The algorithm iteratively performs the following steps:
\begin{enumerate}
    \item Noising: Add noise from $x_0$ to $x_k$ for some $k \in [0,T]$. 
    \item Inference-time alignment: Use inference-time technique to transition back from $x_k$ to $x_0$. 
    \item Selection: Retain only the designs that achieve high target rewards and satisfy specific constraints. For example, constraints could include an edit distance threshold relative to predefined seed sequences.
\end{enumerate}

As a special case, when $k=T+1$, the algorithm reduces to the previously introduced inference-time techniques. However, in this scenario, the algorithm may perform poorly if the constraints are difficult to satisfy. By selecting a moderate $k$ closer to $0$, it is possible to generate samples that better adhere to the given constraints.

{
\subsection{Connection with Evolutionary Algorithms}

Interestingly, we point out that \pref{alg:decoding4} can be viewed as a variant of evolutionary algorithms \citep{branke2008multiobjective}. Recall that a standard evolutionary algorithm iteratively follows the two steps below:
\begin{itemize}
    \item \textbf{Mutation}: Generate candidate sequences based on the current designs (e.g., by introducing noise). 
    \item \textbf{Selection}:  Retain only those designs that achieve high target rewards and satisfy the given constraints. For example, constraints could include an edit distance threshold relative to predefined seed sequences.
\end{itemize}

In our context, it is evident that the mutation step corresponds to the first stage of our algorithm. However, \pref{alg:decoding4} adopts a more strategic mutation approach by leveraging inference-time techniques instead of random mutations to generate candidate sequences. This incorporation of inference-time techniques significantly enhances the efficient exploration of natural-like designs.

} 

{ \section{Comparison with Inference-Time Techniques in Language Models}\label{sec:autoregresstive}

{ This section examines the connections and distinctions between guidance methods in diffusion models and language models. We begin by comparing diffusion models with autoregressive language models, such as GPT \citep{brown2020language}, and then proceed to discuss comparisons with masked language models, such as BERT \citep{kenton2019bert}
. The key message is as follows. 

\begin{AIbox}{Key Message}
  \begin{itemize}
    
\item Most of the guidance methods discussed so far can be directly applied to autoregressive models. However, in autoregressive models, training-free approaches to approximate value functions cannot be utilized due to the lack of forward policies. Additionally, continuous-time formulations play a minimal role within this context.

\item Recall that masked diffusion models can be viewed as hierarchical versions of masked language models. Compared to standard masked language models, masked diffusion models are less susceptible to distributional shifts between training and inference phases. By leveraging this connection, we can develop guidance methods in masked language models by drawing analogies to diffusion models. 
  \end{itemize}
\end{AIbox}
} 

\subsection{Inference-Time Alignment Technique in Autoregressive Models}

In the context of autoregressive pre-trained models, which are prevalent in language models, approaches analogous to derivative-free guidance (\pref{sec:derivative_free}) and derivative-based guidance (\pref{sec:derivative}) have been explored. These methods are summarized in Table~\ref{tab:tab_correpsponding}. While we acknowledge the significant similarities between autoregressive models and diffusion models, we highlight key properties unique to diffusion models that are not accessible in autoregressive models, as discussed below.

\begin{table}[!th]
    \centering
    \caption{Examples of corresponding methods between diffusion models and autoregressive models}
    \begin{tabular}{c|c|p{0.45\textwidth}} 
Algorithm   &   Diffusion models     &     Autoregressive models \\ \midrule 
    SMC-based guidance &  \pref{sec:SMC}  & \citet{zhao2024probabilistic,lew2023sequential} \\ 
    Value-based sampling & \pref{sec:beam}  &  \citet{yang2021fudge,deng2023reward,mudgal2023controlled,han2024value,khanov2024args} \\
    Derivative-based guidance & \pref{sec:derivative},\,\ref{sec:derivative_discrete}  & \citet{dathathri2019plug,qin2022cold} (Plug-in-play approach)    \\ 
        MCTS   & \pref{sec:MCTS} & \citet{leblond2021machine,feng2023alphazero}  
    \\ \bottomrule 
    \end{tabular}
    \label{tab:tab_correpsponding}
\end{table}

\subsubsection{Properties Leveraged in Diffusion Models}

We highlight three fundamental properties of diffusion models that distinguish them from autoregressive models in the construction of inference-time algorithms.

\paragraph{Existence of Forward Noising Processes.}

The training of diffusion models employs a forward process (noising process) from $x_0$ to $x_t$, which proves useful not only during training but also in post-training techniques like policy distillation (as discussed in \pref{sec:policy_distil}) and pure distillation methods \citep{salimans2022progressive}.

\paragraph{Training-Free Value Functions.}

Due to the abovementioned forward noising processes, in diffusion models, a one-step denoising mapping from $x_t$ to $x_0$ (as outlined in Example~\ref{exa:continuous}, \ref{exa:discrete}) simplifies the approximation of value functions due to its training-free nature. In contrast, autoregressive models lack such direct mappings and typically rely on Monte Carlo regression or soft Q-learning (\pref{sec:method_value}) to construct value functions, adding significant complexity \citep{han2024value, mudgal2023controlled}.

\paragraph{Utility of Continuous-Time Formulation.}

In autoregressive models, plug-and-play methods add gradients of classifiers or rewards at inference time \citep{dathathri2019plug, qin2022cold}. While classifier guidance in diffusion models serves a similar purpose, the continuous-time formulation discussed in \pref{sec:derivative} or \pref{sec:derivative_discrete} provides a more formal framework for diffusion models. This formulation is critical in constructing several algorithms and foundations in derivative-based guidance.  

\subsection{Inference-Time Alignment in Masked Language Models}

{ In masked language models, we create inputs with a certain probability of masking and aim to train neural networks (encoders) to demask, i.e., predict the original inputs from the masked ones. Compared to autoregressive language models, masked models are known for their strengths in representation learning, which has led to their widespread use in biology. While lower masking rates are typically employed to capture good representations by understanding global contexts, modern masked language models in biology, such as ESM3 \citep{hayes2024simulating} or masked autoencoders for images \citep{he2022masked} use more aggressive masking rates to enhance their generative capabilities as well.

As noted in Example~\ref{exa:discrete}, masked language models are closely related to masked diffusion models. We will explore these similarities and differences in detail. Then, we discuss alignment methods in masked language models.

\subsubsection{Similarities and Differences Compared to Masked Diffusion Models} 

Both training approaches between masked diffusion models and masked language models are fundamentally similar, although the masking rate in masked language models is typically much lower. The key distinction lies in the decoding approach as follows. 

\paragraph{Masked Diffusion Models.} In masked diffusion models, the decoding process is mathematically structured to ensure that the marginal distributions of the ``noised distributions'' ( induced by $q_t$ from $t=0$ to $t=T$ in Example~\ref{exa:discrete})  and the ``denoised distributions'' (induced by $p_t$ from $t=T$ to $t=0$ in Example~\ref{exa:discrete}) are identical. Hence, since this implies that the training distribution and test distribution at inference time are similar, the learned encoder (neural networks mapping $x_t$ to $x_0$) avoids distributional shifts. Relatedly, the likelihood of a sample, which plays a crucial role in biological applications such as variant effect prediction \citep{livesey2023advancing} or as a key measure to characterize fitness (naturalness) \citep{hie2024efficient}, can be formally computed using the ELBO bound. This bound offers a formal guarantee in the sense that, in the continuous-time limit as $T$ approaches infinity and with an ideal nonparametric neural network, it is tightly achieved. More specifically, when $T\to \infty$, recalling \eqref{eq:loss_discrete}, it is approximated by 
\begin{align*}
    \int_{t=0}^{t=1} \EE_{x_t \sim q_t(\cdot\mid x_0)}\left [ \frac{\bar \alpha'_t}{1-\bar \alpha_t}\rI(x_t = \textbf{Mask})\log \langle x_0,\log \hat x_0(x_t)\rangle \right ]dt. 
\end{align*}

\paragraph{Masked Language Models.} In masked language models, specific positions to unmask must be defined, with various decoding approaches available, such as confidence-based decoding, left-to-right decoding, and random decoding. While these decoding methods bear certain similarities to masked diffusion models, in which inputs are progressively masked, no universally prescribed approach exists.

Even when a particular decoding method is selected, distributional shift remains a potential issue, as the learned encoder may not accurately reflect the distribution induced by the chosen decoding strategy (i.e., a mismatch between the training and inference-time distributions). Additionally, ``formally'' calculating the likelihood is challenging, as it is dependent on the decoding method. Given a sequence with $x^{1:L}$, standard approximation techniques include pseudo-likelihood estimation \citep{miao2019cgmh}: 
\begin{align*}
\sum_{i=1}^L  \log p(x^{i}| (x^{1:L} \setminus x^{i}) ), 
\end{align*}
where $(x^{1:L} \setminus x^{i})$ denotes the sequence $x^{1:L}$ with the $i$-th token replaced by a masked token. Another way is to use Monte Carlo sampling across multiple decoding strategies:
{\begin{align*}
  \frac{1}{|\mathcal{G}|}\sum_{\sigma \in \Gcal} \sum_{i=1}^L \log p(x^{\sigma(1)\sigma(2)\cdots\sigma(i)}|x^{\sigma(1)\sigma(2)\cdots\sigma(i-1)}), 
\end{align*}
} 
where $\mathcal{G}$ is a set that consist of permutation among $L$ tokens. 
However, these methods lack formal guarantees or could easily suffer from distributional shift.

\subsubsection{Adaptation of Alignment Methods}

Various decoding methods have been extensively explored in masked language models \citep{miao2019cgmh,wang2019bert}. However, to the best of the author’s knowledge, inference-time alignment methods have received less attention, partly due to the recent success of autoregressive models over masked language models in generative tasks. Here, we illustrate how the alignment methods discussed for diffusion models can be readily adapted for masked language models.

Suppose we aim to decode from $x_T$ (fully masked) to $x_0$ (non-masked), where each 
$x$ belongs to the space with size $|K|^L$, with $K$ as the vocabulary size and $L$ as the token length. For $x_t\,(t\in[0,T])$, several tokens remain masked. Drawing an analogy from diffusion models, we now explain the formulation of the pre-trained policy and value function in masked language models.

\begin{itemize}
    \item \textbf{Pre-Trained Denoising Policy}: Suppose we are now at $x_{t}$. At this point, we need to decide which position to unmask. We can use any standard method, such as confidence-based selection, random selection, or left-to-right selection (similar to autoregressive models). After selecting the position, we determine the most suitable token using the encoder’s output using top-K-sampling, for example. We refer to this as the pre-trained policy $p^{\pre}_{t-1}(\cdot \mid x_t)$ mapping from $|K|^{L}$ to $\Delta(|K|^{L} ) $ in masked language models.    
    \item \textbf{Value Function}: 
Suppose we are now at $x_t$. In masked language models, we can unmask from $x_t$ to $x_0$ in a single step. Denoting this as $\hat x_0(x_t)$, we define $r(\hat x_0(x_t))$ as the value function. This approach is analogous to defining the value function as in the posterior mean approximation, as discussed in \pref{sec:posterior_mean}. 
\end{itemize} 

We are now prepared to discuss alignment methods. After defining pre-trained policies and value functions, a natural strategy is to approximate the policy $$p^{\star}_{t-1}(x_{t-1} \mid x_t)\propto \exp(r(\hat x_0(x_{t-1}))/\alpha) p^{\pre}_{t-1}(x_{t-1} \mid x_t)$$ to sample from $x_{t-1}$ at the next time step. Any of the previously discussed off-the-shelf strategies, such as SMC-based guidance in \pref{sec:SMC} or value-based sampling in \pref{sec:SVDD}, can be applied here.

}

\section{Combining Fine-Tuning with Inference-Time Techniques}\label{sec:fine_tuning} 

\begin{figure}[!ht]
\centering
\includegraphics[width=0.9\textwidth]{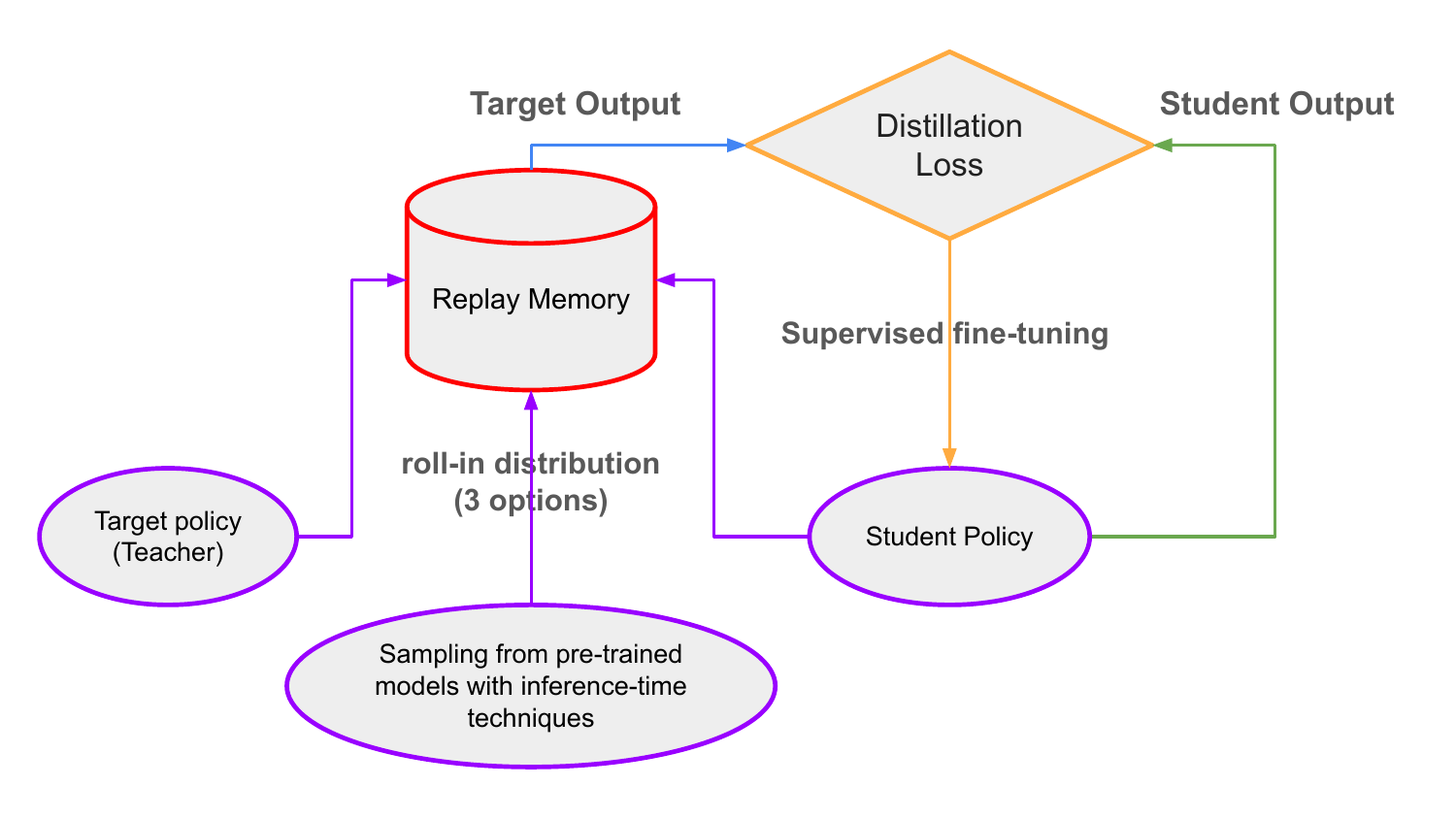}
  \caption{Visualization of policy distillation that leverages inference-time techniques in fine-tuning.
  }
\label{fig:distil}
\end{figure}

While the inference-time techniques discussed so far have proven effective, a potential bottleneck lies in the extended inference time required to generate samples. For instance, in derivative-free guidance (\pref{sec:derivative_free}), value functions must be evaluated at each time step. Similarly, in derivative-based guidance (\pref{sec:derivative}), the derivatives of value functions need to be computed.

To mitigate this issue, we outline fine-tuning strategies to accelerate inference time. Our primary approach involves \emph{policy distillation}, a widely adopted technique in RL \citep{rusu2015policy,czarnecki2019distilling}. We also emphasize its connection to RL-based fine-tuning in diffusion models, which is summarized in \citet{uehara2024understanding}. Finally, we briefly review methods for minimizing inference time in the context of pure diffusion models (i.e., without alignment considerations), which can be applied following policy distillation.

Before delving into the details of policy distillation, we first emphasize the differences between inference-time techniques and standard post-training methods in diffusion models, such as classifier-free fine-tuning and RL-based fine-tuning.

\subsection{Comparison with Standard Post-Training Methods}

We compare inference-time techniques with several alternative post-training approaches that achieve the same goal—generating natural-like designs with high functionality.

\subsubsection{Classifier-Free Fine-Tuning}\label{sec:classfier_free}

Classifier-free guidance \citep{ho2022classifier} aims to directly model the target conditional distribution $p^{(\alpha)}(x \mid y)$ by leveraging an unconditional model and adjusting the guidance scale. This method can be applied to fine-tuning when we have a pre-trained model, which characterizes $p^{\pre}\in \Delta(\Xcal)$, and a reward (or classifier) as follows: 

\begin{enumerate} 
\item  Construct a dataset $\mathcal{D} = \{x_i, y_i\}$ where $x \sim p^{\pre}(x)$ and $y = r(x)$ (or $y \sim p(y \mid x)$). If an alternative dataset $\mathcal{D}'$ is available, it can be used to augment the dataset.
\item %
{ Perform classifier-free guidance by fine-tuning the pre-trained model with the constructed dataset. Here, we augment the model with additional parameters to integrate new conditioning $y$, as implemented in ControlNet \citep{zhang2023adding}.} 
\item After fine-tuning, if the objective is conditioning, generate samples conditioned on the target $y$. If the objective is alignment, condition on high values of $y$.
\end{enumerate}

The performance of this approach is highly dependent on the amount of data available in the first step. Specifically, when working with known reward feedback (e.g., inverse problems or non-differentiable simulation-based feedback), this method can be extremely sample-inefficient, as the feedback must be translated into a large dataset that effectively captures the reward signal. Furthermore, this dataset must be transformed into models with additional parameters, which significantly increases computational requirements during both training and inference.

Furthermore, this sample inefficiency becomes more severe when the pre-trained models are conditional diffusion models $p(x \mid c): \mathcal{C} \to \Delta(\mathcal{X})$, and multiple classifiers or rewards $r_1: \mathcal{X} \to \mathbb{R}, \dots, r_M: \mathcal{X} \to \mathbb{R}$ need to be optimized. In such cases, it is necessary to construct pairs of the form $\{c, x, r_1(x), \dots, r_M(x)\}$, rather than separate datasets $\{x, r_i(x)\}$ for each reward function, which requires considerable effort. By contrast, the inference-time techniques reviewed in this work do not involve inefficient data augmentation steps or potentially challenging training processes.

{ Finally, it is worth noting that the inference-time techniques introduced so far could assist in classifier-free fine-tuning during the data augmentation step. In this way, classifier-free fine-tuning and inference-time techniques can be integrated in a complementary manner.} 

\subsubsection{RL-Based Fine-Tuning}

RL-based fine-tuning has been employed to optimize rewards by embedding diffusion models into Markov Decision Processes (MDPs). Seminal works have been proposed using policy gradient methods such as PPO \citep{fan2023dpok,black2023training}. However, given that many important reward functions in computer vision are differentiable, recent state-of-the-art methods focus on variants of direct backpropagation approaches \citep{clark2023directly,prabhudesai2023aligning}.

In molecular design, RL-based fine-tuning presents additional challenges, as most useful feedback is highly non-differentiable. In many cases, policy gradient methods remain necessary over direct backpropagation. However, based on the authors' experience, optimizing such reward functions through policy gradient-based fine-tuning is still difficult, as the landscapes of reward functions are highly complicated. 

In contrast, the inference-time techniques introduced in this review are straightforward to implement, as many of them are not only fine-tuning-free but also training-free. As will be discussed in \pref{sec:foundation}, while both RL-based and inference-time methods ideally aim to reach the same optimal policy, inference-time techniques are generally more stable for alignment. This is because they directly sample from the optimal target policy by guiding individual particles (samples) without altering the underlying diffusion models. On the other hand, RL-based fine-tuning must guide the entire generative model, which encapsulates the information of all generated samples. As a result, RL-based fine-tuning is significantly more challenging. Consequently, inference-time techniques are more effective across various domains, including molecular design, even when reward feedback is highly non-differentiable, as demonstrated in \citet{li2024derivative}.

Lastly, we highlight that inference-time methods can also enhance fine-tuning through policy distillation. We compare this hybrid approach with RL-based fine-tuning in more detail in \pref{sec:fine_tuning}.

\subsection{Policy Distillation}\label{sec:policy_distil}

So far, we have discussed two standard techniques for post-training. In this subsection, we introduce an alternative approach: policy distillation for fine-tuning, leveraging the inference-time techniques presented earlier. The key idea is to fine-tune diffusion models so that the fine-tuned models replicate the trajectories generated by the inference-time technique. Through the iterative application of this process, the fine-tuned models can gradually improve, as illustrated in \pref{fig:distil}.

{ Hereafter, we refer to the inference-time techniques (e.g., SMC-based guidance, value-based importance sampling, and classifier guidance targeting $p^{\star}_{t-1}:\Xcal \to \Delta(\Xcal)$) as teacher policies. The policies being fine-tuned, represented as $p_{t-1}(\cdot \mid \cdot; \theta):\Xcal \to \Delta(\Xcal)$, are referred to as student policies, adopting the terminology from RL \citep{czarnecki2019distilling}.  } 

Algorithms in this section are summarized in the following master formulation.

\begin{AIbox}{Master Formulation}
Introducing a roll-in distribution $u_t \in \Delta(\Xcal)$, and an $f$-divergence between teacher and student policies at time $t$, the policy distillation is formulated as an iterative algorithm using the update: 
\begin{align*}
    \theta^{\mathrm{new}} \leftarrow \theta^{\mathrm{old}}  - \gamma \sum_{t=T+1}^1 \EE_{x_t \sim u_t}[ \nabla_{\theta} 
 f(p^{\star}_{t-1}(\cdot \mid x_t) \|p_{t-1}(\cdot \mid x_t;\theta) )]|_{\theta^{\mathrm{old}}}, 
\end{align*}
where $\gamma$ is a learning rate. 
\end{AIbox}
Hereafter, we elaborate on two critical points.

\subsubsection{Choice of Roll-In Distributions} 

The selection of the roll-in distribution is critical, as optimizing the empirical objective with function approximation ensures low error only within the distribution's support. Potential choices include (1) the teacher policy, (2) the student policy (the one being optimized), and (3) data recycling via forward processes. These distributions can also be mixed. We will explore these options in greater detail.

\paragraph{Teacher Policy.} This approach is often called offline policy distillation, as the roll-in distribution remains fixed throughout fine-tuning. 
Since this corresponds to the target policy, it is always a reasonable choice. However, it could have a distributional shift \citep{ross2010efficient}: during fine-tuning, the model encounters states it has not previously observed yet, which can result in poor performance.

\paragraph{Student Policy.} This approach is often referred to as online policy distillation, as roll-in samples are collected in an online manner. In the RL context, it is employed in the well-known algorithm DAgger \citep{ross2011reduction}. This method mitigates the distributional shift issue mentioned earlier by aligning the training distribution with the distribution induced by the current student policy. In practice, it can be beneficial to mix distributions induced by both teacher and student policies, as demonstrated in \citet{ross2011reduction,sun2017deeply,chang2023learning}.

\paragraph{Data Recycling via Forward Processes.}

In both scenarios described above (using teacher or student policies as roll-in distributions), the roll-in distributions are generated by sampling from $x_T$ to $x_t$. However, employing teacher policies as roll-in distributions can be computationally expensive.

To mitigate this inefficiency, an alternative approach leverages the forward process in pre-trained diffusion models, sampling from $x_0$ to $x_t$, which is computationally faster than sampling from 
 $x_T$ to $x_t$. Specifically, assume we have access to the dataset used to train the pre-trained models or data obtained at time $0$ through inference-time techniques.  Let this dataset be denoted as $\Dcal =\{x^i_0\}_{i=1}^N$.  Using this dataset $\Dcal$, we can generate roll-in distributions by sampling $x_t$ from the forward policies of the pre-trained models, starting from $x_0$. Consequently, the loss function becomes:
\begin{align}
\label{eq:loss_forward_roll_in_distribution}
    \frac{1}{|\Dcal| }\sum_{i=1}^{|\Dcal|} \sum_{t=T+1}^1  \EE_{x_t \sim q_t(\cdot \mid x^i_0) }\left [f(p^{\star}_{t-1}(\cdot \mid x_t) \|p_{t-1}(\cdot \mid x_t;\theta) )\right ], 
\end{align}
where $q_t(\cdot|x^i_0)$ represents the distribution induced by the forward policies of the pre-trained models from $x_0$ to $x_t$.

\subsubsection{Choice of Divergence}

The choice of divergence is critical. Common options include KL divergence and inverse KL divergence.

\paragraph{KL Divergence.} In this case, the gradient is 
\begin{align}\label{eq:KL_appraoch}
 \sum_{t=T+1}^1 \EE_{x_t\sim u_t} \left[ \nabla_{\theta} \KL(p^{\star}_{t-1}(\cdot|x_t) \| p_{t-1}(\cdot \mid x_t;\theta))   \right]|_{\theta^{\mathrm{old}}}. 
\end{align}
This can be further simplified to:
\begin{align*}
 \sum_{t=T+1}^1 \EE_{x_t\sim u_t, x_{t-1}\sim p^{\star}(x_t) } \left[\nabla_{\theta}  \log p_{t-1}(x_{t-1} \mid x_t;\theta)  \right]|_{\theta^{\mathrm{old}}}. 
\end{align*} 
When using marginal distributions induced by teacher policies as roll-in distributions, this approach is particularly intuitive, as it minimizes the KL divergence between the distributions induced by the teacher and student policies. In generative models, KL divergence is commonly employed since it effectively covers the target distribution (i.e., it is less mode-seeking but more conservative).

\paragraph{Inverse KL Divergence.}

In this case, the gradient is 
\begin{align}
    \sum_{t=T+1}^1 \EE_{x_t\sim u_t} \left[\nabla_{\theta} \KL(p_{t-1}(\cdot \mid x_t;\theta) \| p^{\star}_{t-1}(\cdot|x_t))   \right]|_{\theta^{\mathrm{old}}}.  \label{eq:inverse_KL}
\end{align}
Substituting the form of the optimal policy $p^{\star}_{t-1}$, it simplifies to:
\begin{align}
  \sum_{t=T}^0  \EE_{x_t \sim u_t} \left[ \nabla_{\theta}\EE_{x_{t-1}\sim p_{t-1}(x_t;\theta) } \left [ \log p_{t-1}(x_{t-1}|x_t;\theta) -\log p^{\pre}_{t-1}(x_{t-1}\mid x_t) -\frac{v_{t-1}(x_{t-1})}{\alpha} + \frac{v_{t}(x_{t})}{\alpha}    |x_t \right] \right]. \label{eq:all_form}
\end{align}

This method is inherently more mode-seeking. While this property can be disadvantageous when the goal is to cover diverse regions, it becomes beneficial when the primary focus is on optimization without prioritizing diversity.

\subsection{Relation to RL-Based Fine-Tuning}

In this section, we establish the connection between the policy distillation approach discussed earlier and existing RL-based fine-tuning methods, which are summarized in \citet{uehara2024understanding}. 

\subsubsection{Value-Weighted MLE (Reward-Weighted MLE)}

First, we relate the KL-based distillation approach to value-weighted maximum likelihood estimation (MLE). Recall that $p^{\star}_{t-1}$ is a value-tilted policy proportional to $p^{\pre}_{t-1}(\cdot \mid x_{t-1})\exp(v_{t-1}(\cdot))$. When the divergence function $f$ is the KL divergence, the gradient formulation (\ref{eq:KL_appraoch}) becomes:
\begin{align*}
      \sum_{t=T+1}^1 \EE_{x_t \sim u_t}\left [ \nabla_{\theta} \EE_{ x_{t-1}\sim p^{\pre}(x_t)}\left [\exp\left ( \frac{v_{t-1}(x_{t-1})}{\alpha} \right)\log p_{t-1}(x_{t-1}|x_t;\theta) \right] \right ]. 
\end{align*}

This formulation is known in the RL literature as value-weighted MLE \citep{peng2019advantage,peters2010relative}. In the context of fine-tuning diffusion models, it reduces to the objective function defined in \citet[Algorithm 3]{uehara2024understanding}. 

\subsubsection{Path Consistency Learning (loss used in Gflownets)}

Next, recall the soft-Bellman equation:
\begin{align*}
    \int p^{\pre}_{t-1}(x|x_{t})\exp(v_{t-1}(x)/\alpha)dx = \exp(v_{t}(x_t)/\alpha). 
\end{align*}
Then, by taking the logarithm of the optimal policy
\begin{align*}
    p^{\pre}_{t-1}(\cdot|x_{t})= \exp(v_{t-1}(\cdot)/\alpha)p^{\pre}_{t-1}(x|x_{t})/ \exp(v_{t}(x_t)/\alpha), 
\end{align*}
this reduces to 
\begin{align*}
     \log p^{\star}_{t-1}(x_{t-1}|x_t) -\log p^{\pre}_{t-1}(x_{t-1}\mid x_t) -\frac{v_{t-1}(x_{t-1})}{\alpha} + \frac{v_{t}(x_{t})}{\alpha}=0. 
\end{align*}
From the above equation, the optimal policy is estimated by minimizing the following objective: 
\begin{align*}
     \sum_{t=T+1}^1  \EE_{(x_{t-1},x_t) \sim u_t} \left[\left \| \log p_{t-1}(x_{t-1}|x_t;\theta) -\log p^{\pre}_{t-1}(x_{t-1}\mid x_t) -\frac{v_{t-1}(x_{t-1})}{\alpha} + \frac{v_{t}(x_{t})}{\alpha}   \right \|^2 \right]. 
\end{align*}
This objective aligns with the path-consistency RL objective commonly discussed in the RL literature \citep{nachum2017bridging}. In RL-based fine-tuning, it reduces to the objective defined in \citet[Algorithm 5]{uehara2024understanding}. Note several improved variants have also been proposed in \citet{rector2024steering,venkatraman2024amortizing}.

The above is closely related to the inverse KL divergence objective \eqref{eq:all_form}. However, PCL is generally regarded as more stable, as the expectation in \eqref{eq:all_form} depends on the parameter being optimized.

\subsubsection{PPO and Direct Backpropagation}

In the RL-based fine-tuning literature~\citep{fan2023dpok,black2023training}, the standard objective is:
\begin{align}\label{eq:PPO}
   \argmax_{\theta} \EE_{\{p^{\theta}_t\}_{t=T+1}^1} \left[r(x_0)- \alpha \sum_{t=T+1}^1 \KL(p_{t-1}(\cdot|x_t;\theta) \| p^{\pre}_{t-1}(\cdot|x_t))   \right], 
\end{align}
where the expectation is taken with respect to $p^{\theta}$. As demonstrated in \citet[Theorem 1]{uehara2024bridging}, the policy that maximizes the above (in the absence of optimization or function approximation errors) is equal to the optimal policy we aim to target during inference, i.e., $p^{\star}_{t-1}$ in \eqref{eq:optimal_policy}. 

While solving the optimization problem in \eqref{eq:PPO} is non-trivial in practice, policy gradient methods (or their variants) can be employed \citep{black2023training,fan2023dpok}. Then, ignoring KL terms to simplify the notation here, the optimization step becomes 
\begin{align}\label{eq:PPO2}
    \theta^{\mathrm{old}}  +  \gamma \EE_{\{p^{\theta^{\mathrm{old}}}_t\}_{t=T+1}^1} \left[r(x_0) \sum_{t=T+1}^1 \frac{\nabla_{\theta} p_{t-1}(\cdot \mid x_{t};\theta )}{p_{t-1}(\cdot \mid x_{t};\theta^{\mathrm{old}})}    \right]|_{\theta^{\mathrm{old}}}.
\end{align}
Furthermore, when the reward function is differentiable, direct backpropagation becomes a feasible approach \citep{clark2023directly,prabhudesai2023aligning}. 

It is worthwhile to note the above objective function in \eqref{eq:PPO} is essentially equivalent to the inverse KL divergence: 
\begin{align*}
   \sum_{t=T+1}^1  \EE_{\{p^{\theta}_t\}_{t=T+1}^1}[\mathrm{KL}(p_{t-1}(\cdot\mid x_{t-1};\theta)\|p^{\star}_{t-1}(\cdot\mid x_{t-1}))  ]. 
\end{align*}
This is seen by 
{\begin{align*}
    & - \alpha \sum_{t=T+1}^1 \EE_{\{p^{\theta}_t\}_{t=T+1}^1}[\mathrm{KL}(p_{t-1}(\cdot\mid x_{t-1};\theta)\|p^{\star}_{t-1}(\cdot\mid x_{t-1}))  ]\\
    &=- \alpha \sum_{t=T+1}^1 \EE_{\{p^{\theta}_t\}_{t=T+1}^1}\left [  \log p_{t-1}(x_{t-1}|x_t;\theta) -\log p^{\pre}_{t-1}(x_{t-1}\mid x_t) - v_{t-1}(x_{t-1}) + v_{t}(x_{t})  \right]  \\
        &= \EE_{\{p^{\theta}_t\}_{t=T+1}^1} \left[r(x_0) - \alpha \sum_{t=T+1}^1 \KL(p_{t-1}(\cdot|x_t;\theta) \| p^{\pre}_{t-1}(\cdot|x_t))   \right]. 
\end{align*} } 
Thus, RL-based fine-tuning is considered to be similar to the policy distillation approach using the inverse KL divergence. 

However, the distillation approach using inverse KL, as discussed in \eqref{eq:inverse_KL}, uses the estimation of value functions, leading to substantially different practical behavior, while the PPO/direct backpropagation approaches do not estimate value functions explicitly. 

\subsection{Differences between Policy Distillation and RL-Based Fine-Tuning}
\label{sec:compare_distill}

Here, we briefly compare two approaches: policy distillation and RL-based fine-tuning. For clarity, by policy distillation, we refer to the simplest KL-based approach in \eqref{eq:KL_appraoch}, and by RL-based fine-tuning, we refer to the PPO approach \citep{black2023training,fan2023dpok}, which optimizes the objective function in \eqref{eq:PPO} with PPO \citep{schulman2017proximal}.

\paragraph{Advantages of Policy Distillation over PPO.} 

The primary advantage lies in the stability of the fine-tuning process. Policy distillation using KL divergence is equivalent to supervised learning toward target optimal policies, ensuring stable fine-tuning since the target remains fixed. {Additionally, policy distillation can function as an offline algorithm, with roll-in distributions independent of current policies (student policies) and could be technically any roll-in distributions.}

In contrast, PPO introduces instability due to the need for continuously updating roll-in distributions and optimizing a dynamic objective function involving ratios relative to the evolving policies being trained. Although PPO incorporates conservative updates with KL constraints (as seen in TRPO \citep{schulman2015trust} and CPI \citep{kakade2002approximately}) to mitigate this instability, ensuring stable updates remains challenging in practice, often resulting in convergence to suboptimal local minima before reaching optimal policies. Furthermore, while related methods, such as direct backpropagation \citep{clark2023directly,prabhudesai2023aligning}, can address some of these optimization challenges, they require differentiable models, which are difficult to construct in molecular design.

A second advantage is policy distillation’s robustness against reward over-optimization. This issue arises when generative models over-exploit learned reward functions, resulting in out-of-distribution samples that achieve low genuine rewards. Policy distillation inherently avoids this issue since roll-in samples are generated using inference-time techniques, i.e., essentially modified pre-trained models, that are more likely to stay on the manifold, reflecting natural-like designs (e.g., natural image spaces in computer vision or chemical/biological spaces in molecular design).

In contrast, PPO’s dynamic roll-in distributions can easily deviate from these natural manifolds. Although techniques such as KL penalization against deviations from pre-trained policies, as outlined in \eqref{eq:PPO}, and the use of conservative reward models \citep{uehara2024bridging} can help mitigate this issue, the non-conservative nature of the approach may still pose challenges.

\subsection{Further Extensions with ``Distillation''}

After distilling optimal policies, we can apply additional standard ``distillation'' techniques in diffusion models. Notably, unlike policy distillation, the typical objective of these ``distillation'' methods is to reduce inference time while preserving the naturalness of the generated samples, without involving reward maximization.

\paragraph{Distillation.}

The focus of distillation in diffusion models is to reduce the number of discretization steps while maintaining the quality of the generated samples \citep{salimans2022progressive, sun2023accelerating, meng2023distillation}. For example, a widely adopted approach is progressive distillation \citep{salimans2022progressive}, which iteratively halves the number of discretization steps with each stage of the distillation process.

\paragraph{Consistency Distillation/Training.}

Consistency training aims to ensure that the model's predictions remain stable across multiple timesteps during the reverse diffusion process. The primary motivation is to minimize the number of reverse steps required without compromising the quality of the generated samples. For further details, refer to \citet{song2023consistency, li2024towards, ding2023consistency}.

\section{More Related Works} \label{sec:related_works}

We have covered various aspects of inference-time techniques in diffusion models. In this section, we finally outline additional related topics that were not the primary focus of our discussion.

\subsection{Walk-Jump Sampling for Protein Design}\label{sec:refinement}

Walk-Jump Sampling \citep{frey2023protein,saremi2019neural} is an algorithm designed to achieve high reward values while preserving the naturalness of the generated designs, making it particularly effective for antibody design. 

The intuition behind walk-jump sampling can be summarized as follows. Similar to our objective in \pref{sec:goal},  the goal is to sample from the distribution $\exp(r(x))/\alpha )p^{\pre}(x)$. One approach to achieve this is through Langevin Monte Carlo (LMC), also known as the Metropolis-adjusted Langevin algorithm (MALA): sampling using the following update: 
{ \begin{align}\label{eq:MALA}
    y_{k} \leftarrow  y_{k-1} +  \beta\{\nabla r(y_{k-1})/\alpha  + \nabla_{x} \log p^{\pre}(x)|_{y_{k-1}} \} + \sqrt{2\beta}\epsilon_k,\quad \epsilon_k\sim \Ncal(0,1).
\end{align}
} 
 However, the score function $\nabla_{x} \log p^{\pre}(x)$ is unknown. A common method for estimating the score is through denoising score matching \citep{vincent2011connection}: 
\begin{align*}
   \hat \theta= \argmin_{\theta} \EE_{x\sim p^{\pre},\tilde x \sim \Ncal(x,\sigma^2 I)}\left[ \left \|\frac{x-\tilde x}{\sigma^2} - s(\tilde x;\theta) \right \|^2 \right ]. 
\end{align*}
By plugging this into \eqref{eq:MALA}, the LMC algorithm is an iterative algorithm using the following update: 
\begin{align}\label{eq:MALA2}
    y_{k} \leftarrow y_{k-1}  + \beta \{ \underbrace{ \nabla r(y_{k-1})/\alpha}_{\textbf{Walk}} +  \underbrace{s(y_{k-1};\hat \theta)}_{\textbf{Jump} } \} + \sqrt{2\beta}\epsilon_k. 
\end{align}
{ Building on this intuition, walk-in jump sampling is defined as an algorithm that iteratively performs a ``walk'' (adding the gradient of the reward) followed by a ``jump'' (adding a score) as described above.}

Importantly, walk-in jump sampling can be viewed as a variant of classifier guidance, where a single noise level is employed, as follows. For this purpose, recall that while the diffusion model can be framed in terms of variational inference \citep{ho2020denoising} or time-reversal SDEs \citep{song2021score}, another popular derivation involves running MALA with score function estimation at multiple noise levels, known as score-based models (SBMs) \citep{song2019generative}. Thus, walk-jump sampling is similar to to classifier guidance explained in \pref{sec:derivative} where only a single noise level is used.  However, the key difference is that the score and gradient are iteratively updated in an MCMC framework, rather than progressing sequentially from $t = T$ to $t = 0$, as in classifier guidance. For further details, please refer to the original papers.

\subsection{Hallucination Approaches for Protein Design} 

{ The term ``hallucination'' (or its variants) frequently refers to sequential refinement methods in silico protein design. Specifically, this algorithm iteratively refines a sequence based on predefined reward functions. Standard methods for sequence refinement include MCMC \citep{anishchenko2021novo}, evolutionary algorithms \citep{jendrusch2021alphadesign}, and gradient-based algorithms \citep{goverde2023novo,jeliazkov2023esmfold}. In our context, \pref{sec:editing} is closely related. 
Common reward functions include the loss between the predicted structure (using structural prediction models such as AlphaFold or ESMFold \citep{ahdritz2024openfold}) and the target structure, as well as metrics like stability, binding activity, and geometric constraints.

\subsection{Inference-Time Techniques for Inpainting and Linear Inverse Problems.}
Inpainting tasks involve reconstructing or filling in missing or corrupted regions of an image in a manner that appears natural and consistent with the surrounding content. In the context of protein design, a similar challenge is known as motif scaffolding, as mentioned in the introduction. For inpainting tasks, replacement methods (e.g., \citet{song2021score}) are commonly employed. These approaches map the conditioned region (known location) through the forward process and replace it with generated samples at each step of the inference process. While this method is somewhat heuristic, a more refined version incorporating SMC was later proposed by \citet{trippe2022diffusion}, demonstrating success in motif scaffolding.

Additionally, for related tasks like linear inverse problems, more specialized algorithms have recently been introduced \citep{song2023pseudoinverse,chung2022improving,kawar2022denoising}. Although these methods may not be as broadly applicable as those discussed here, they offer superior performance in specific, constrained scenarios.

\subsection{Speculative Decoding.} Speculative decoding is an inference-time technique used in NLP to accelerate the generation process \citep{leviathan2023fast}. The key idea is to generate multiple candidate tokens simultaneously using a smaller, faster model and validate them with a larger, more powerful model. As models grow larger in protein design, adopting similar techniques could offer substantial benefits in accelerating inference in this field as well.

\section*{Acknowledgement}

The authors would like to thank Amy Wang, Francisco Vargas, Christian A. Naesseth for valuable discussions and feedback.

\bibliographystyle{chicago}
\bibliography{rl,rl2}

\end{document}